\pdfoutput=1
\PassOptionsToPackage{table}{xcolor}
\RequirePackage{fix-cm}
\documentclass[]{xiaomiev}
\let\gfxbyname\includegraphics

\microtypesetup{expansion=false}
\DeclareFontShape{T1}{xiaomiev}{b}{sc}{<-> s * [1] xiaomiev}{}
\usepackage{xspace}
\usepackage{mathtools,amsmath,amssymb}
\usepackage{enumitem}
\usepackage{needspace}
\usepackage{fontawesome5}   %
\usepackage{tabularx}
\usepackage{colortbl}
\usepackage{etoolbox}

\definecolor{hxOrange}{HTML}{ED722E}
\definecolor{hxNavy}{HTML}{1B262C}
\definecolor{hxBlue}{HTML}{0F4C75}
\definecolor{hxTeal}{HTML}{3282B8}
\definecolor{hxSky}{HTML}{BBE1FA}
\definecolor{miBest}{HTML}{FF7E00}
\definecolor{miSecond}{HTML}{FFBE80}
\definecolor{cPR}{HTML}{718FBF}
\definecolor{cP}{HTML}{D891A2}
\definecolor{cE}{HTML}{8DB879}
\definecolor{cF}{HTML}{D8B85E}
\definecolor{cN}{HTML}{A1A5AA}
\definecolor{cMZ}{HTML}{7E8490}
\colorlet{cBox}{miBest!10}
\definecolor{cUp}{HTML}{527D40}
\definecolor{cDn}{HTML}{A45E68}
\definecolor{paperInk}{HTML}{252A31}
\colorlet{paperRule}{black}
\definecolor{paperGrid}{HTML}{E4E7EB}

\newcommand{\papertablefont}{\footnotesize}

\newcolumntype{R}{>{\raggedleft\arraybackslash}X}
\AtBeginEnvironment{table}{%
  \papertablefont\setlength{\tabcolsep}{5pt}%
  \renewcommand{\arraystretch}{1.05}%
  \arrayrulecolor{black}}

\setlist[itemize]{leftmargin=15pt}
\usepackage[T1]{fontenc}
\usepackage[utf8]{inputenc}
\graphicspath{{figures/}}
\usepackage{algorithm}
\usepackage{algpseudocode}
\floatstyle{ruled}
\newfloat{humanprocedure}{tbp}{lop}
\floatname{humanprocedure}{Procedure}
\algrenewcommand{\algorithmicrequire}{\textbf{Input:}}
\algrenewcommand{\algorithmicensure}{\textbf{Output:}}

\newcommand{\mlogo}[1]{\raisebox{-.17em}{\gfxbyname[height=.92em]{figures/logos/#1.png}}\hspace{.32em}}

\DeclareTextFontCommand{\textbf}{\bfseries\color{black}}
\titleformat*{\paragraph}{\normalfont\normalsize\bfseries\color{black}}
\pretocmd{\section}{\Needspace{8\baselineskip}}{}{}
\pretocmd{\subsection}{\Needspace{5\baselineskip}}{}{}

\title{{\fontsize{16}{20}\selectfont Beyond Task Success: A Tool-Use Agent Efficiency Benchmark}}
\DeclareRobustCommand{\centeredlogo}[2]{\raisebox{-.5\height}{\gfxbyname[#1]{figures/logos/inst/#2}}}
\DeclareRobustCommand{\numberedlogo}[3]{\textsuperscript{\normalfont\scriptsize #1}\kern1.5pt\centeredlogo{#2}{#3}}
\newcommand{\institutionlogos}{%
  \noindent\makebox[\linewidth]{\hfil
    \numberedlogo{1}{width=110pt,height=32pt,keepaspectratio}{iie-university-official.png}\hfil
    \numberedlogo{2}{width=75pt,height=31pt,keepaspectratio}{wuerzburg-official.png}\hfil
    \numberedlogo{3}{width=78pt,height=22pt,keepaspectratio}{Xiaomi_logo_horizontal_2021.pdf}\hfil
    \numberedlogo{4}{width=68pt,height=31pt,keepaspectratio}{brown.pdf}\hfil}\par\vspace{6pt}%
  \noindent\makebox[\linewidth]{\hfil
    \numberedlogo{5}{width=46pt,height=33pt,keepaspectratio}{cmu-university-official.png}\hfil
    \numberedlogo{6}{width=88pt,height=29pt,keepaspectratio}{uwa.png}\hfil
    \numberedlogo{7}{width=66pt,height=29pt,keepaspectratio}{kit.pdf}\hfil
    \numberedlogo{8}{width=120pt,height=25pt,keepaspectratio}{cuhk-university-official.png}\hfil}\par\vspace{6pt}%
  \noindent\makebox[\linewidth]{\hfil
    \numberedlogo{9}{width=72pt,height=32pt,keepaspectratio}{nus.png}\hfil
    \numberedlogo{10}{width=100pt,height=28pt,keepaspectratio}{hust.png}\hfil
    \numberedlogo{11}{width=70pt,height=33pt,keepaspectratio}{uw.png}\hfil
    \numberedlogo{12}{width=118pt,height=25pt,keepaspectratio}{columbia.pdf}\hfil}%
}
\renewcommand{\authorformat}[2][]{{\fontsize{10}{13}\selectfont\rmfamily #2}}
\author{%
\mbox{Yu Liu\textsuperscript{1,*}}\quad
\mbox{Yuedong Tan\textsuperscript{2,*,\ddag}}\quad
\mbox{Zhilin Liu\textsuperscript{3,*}}\quad
\mbox{Zhiwei Yang\textsuperscript{1,*}}\quad
\mbox{Zheyuan Deng\textsuperscript{4}}\quad
\mbox{Hanqi Feng\textsuperscript{5}}\quad
\mbox{Wenxiao Zhang\textsuperscript{6}}\quad
\mbox{Kunyu Peng\textsuperscript{7}}\quad
\mbox{Radu Timofte\textsuperscript{2}}\quad
\mbox{Youxing LI\textsuperscript{8}}\quad
\mbox{Zixiang Peng\textsuperscript{1}}\quad
\mbox{Di Wen\textsuperscript{7}}\quad
\mbox{Junwei Zheng\textsuperscript{7}}\quad
\mbox{Ruiping Liu\textsuperscript{7}}\quad
\mbox{Yufan Chen\textsuperscript{7}}\quad
\mbox{Guobin Zhao\textsuperscript{9,\dag}}\quad
\mbox{Minheng Wu\textsuperscript{7}}\quad
\mbox{Wenhao Guo\textsuperscript{7}}\quad
\mbox{Zhiyuan Ma\textsuperscript{10,\dag}}\quad
\mbox{Weitong Kong\textsuperscript{7}}\quad
\mbox{Ruodi Zhang\textsuperscript{7}}\quad
\mbox{Haiwen Sun\textsuperscript{7}}\quad
\mbox{Jiazhen Meng\textsuperscript{7}}\quad
\mbox{Qingshuo Lyu\textsuperscript{1}}\quad
\mbox{Yanjing Li\textsuperscript{11}}\quad
\mbox{Tingli Qian\textsuperscript{11}}\quad
\mbox{Xuan Qi\textsuperscript{11}}\quad
\mbox{Chenhao Zhang\textsuperscript{11}}\quad
\mbox{Xiuwen Xi\textsuperscript{3}}\quad
\mbox{Shaojie Zhang\textsuperscript{3}}\quad
\mbox{Pei Fu\textsuperscript{3,\dag}}\quad
\mbox{Jin B. Hong\textsuperscript{6}}\quad
\mbox{Yanbing Liu\textsuperscript{1}}\quad
\mbox{Jian Luan\textsuperscript{3}}\quad
\mbox{Barnabas Poczos\textsuperscript{5}}\quad
\mbox{Li Erran Li\textsuperscript{12}}\par\vspace{4pt}
{\normalfont\fontsize{8}{10}\selectfont
\textsuperscript{*}Equal contribution.\quad
\textsuperscript{\ddag}Project leader.\quad
\textsuperscript{\dag}Corresponding authors.\\[8pt]
\institutionlogos}}

\abstract{
An agent auditing an expense claim can inspect several receipts together or repeatedly alternate between reading evidence and performing calculations. Both approaches may reach the correct decision while revealing different ways of organizing external work. Understanding tool agents requires relating these behaviors to what their interactions accomplish. We introduce \textbf{TAE-Bench}, a multimodal benchmark with 191 tasks for evaluating final correctness alongside evidence acquisition in long-horizon tasks. Models operate through a shared base harness and tool environment, controlling variation across agent frameworks. Each task has an evidence graph linking facts, intermediate results, and required states, with executable predicates recognizing evidence across alternative tool sequences. Our evaluation of 27 frontier models separates capability, family-associated tool preferences, and execution organization. Models from the same company use more similar tool combinations on jointly solved tasks, while variants within a family can organize calls very differently. Higher-performing variants usually reduce calls and unmatched intermediate work without adopting a universal batching strategy.  TAE-Bench reveals how models acquire evidence and sustain external action beyond final success, supporting further study of tool choice, operation scope, and feedback organization. Constructing TAE-Bench cost approximately US\$12,000, covering task design, evidence-graph annotation, and evaluation. The benchmark tasks and execution environment are publicly available.
}
\definecolor{linkHome}{HTML}{3282B8}   %
\definecolor{linkBoard}{HTML}{D99A00}  %
\protected\def\homelogo{\makebox[1.05em][c]{\raisebox{.02em}{\normalfont\footnotesize\textcolor{linkHome}{\faHome}}}}
\protected\def\boardlogo{\makebox[1.05em][c]{\raisebox{.02em}{\normalfont\footnotesize\textcolor{linkBoard}{\faTrophy}}}}
\protected\def\ghlogo{\raisebox{-.2em}{\includegraphics[height=1.05em]{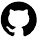}}}
\protected\def\hflogo{\raisebox{-.22em}{\includegraphics[height=1.15em]{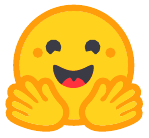}}}
\checkdata[\homelogo\ Project page]{\url{https://huggingface.co/spaces/SeerRay-Lab/TAE-Bench}}
\checkdata[\boardlogo\ Leaderboard]{\url{https://seerray-lab.github.io/TAE-Bench/}}
\checkdata[\ghlogo\ Code]{\url{https://github.com/SeerRay-Lab/TAE-Bench}}
\checkdata[\hflogo\ Data]{\url{https://huggingface.co/datasets/SeerRay-Lab/TAE-Bench}}

\hypersetup{
  pdftitle={Beyond Task Success: A Tool-Use Agent Efficiency Benchmark},
  pdfauthor={Yu Liu; Yuedong Tan; Zhilin Liu; Zhiwei Yang; Zheyuan Deng; Hanqi Feng; Wenxiao Zhang; Kunyu Peng; Radu Timofte; Youxing LI; Zixiang Peng; Di Wen; Junwei Zheng; Ruiping Liu; Yufan Chen; Guobin Zhao; Minheng Wu; Wenhao Guo; Zhiyuan Ma; Weitong Kong; Ruodi Zhang; Haiwen Sun; Jiazhen Meng; Qingshuo Lyu; Yanjing Li; Tingli Qian; Xuan Qi; Chenhao Zhang; Xiuwen Xi; Shaojie Zhang; Pei Fu; Jin B. Hong; Yanbing Liu; Jian Luan; Barnabas Poczos; Li Erran Li},
  pdfsubject={TAE-Bench technical report}
}

\begin{document}
\maketitle\par

\clearpage
\setcounter{tocdepth}{2}
{%
\setlength{\cftbeforesecskip}{0.5pt}\setlength{\cftbeforesubsecskip}{0pt}\small
\tableofcontents}
\clearpage

\begin{figure}[t]
\centering
\includegraphics[width=\linewidth]{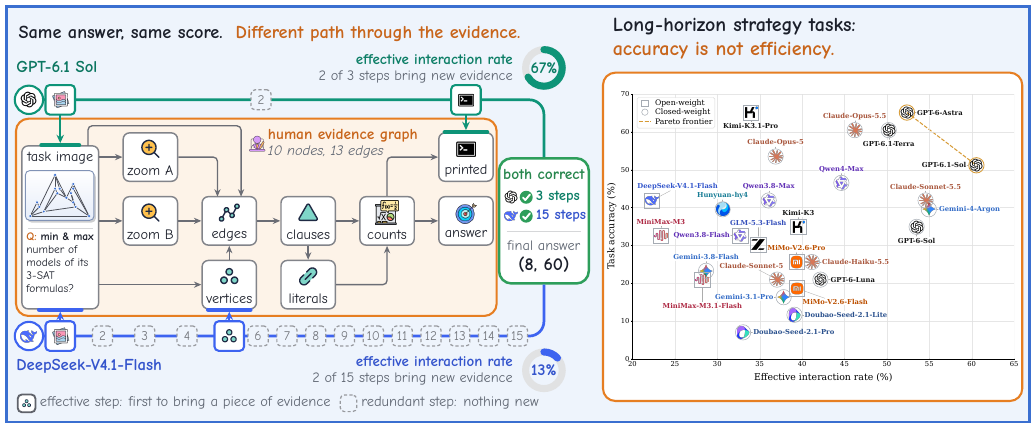}
\caption{\colorbox{red!10}{Same answer, different efficiency.} \emph{Left:} GPT-6.1 Sol and DeepSeek-V4.1-Flash both solve the task, but against its human evidence graph 2 of 3 steps bring new evidence for Sol and only 2 of 15 for DeepSeek. \emph{Right:} on long-horizon strategy tasks, accuracy does not determine the effective interaction rate (EIR). \textbf{Closer to the upper right is better.}}
\label{fig:teaser}
\end{figure}
\section{Introduction}
Language models equipped with tools can retrieve information, manipulate documents, and execute programs through external interfaces~\cite{li2023apibank,qin2024toolllm,patil2025bfcl, qwen2026maxcard,qwen2026qwen4release}. Benchmarks increasingly evaluate whether such agents can complete realistic computer, spreadsheet, and service workflows~\cite{xie2024osworld,ma2024spreadsheetbench,yao2024tau}. As agents undertake longer tasks, understanding how their interactions contribute to completion becomes consequential. An agent must decide what evidence to seek, select tools and parameters, connect intermediate results, and respond to execution feedback. Final correctness establishes whether the requested outcome was achieved, but leaves much of this process unexplained. \textit{Our goal is not to maximize task difficulty, but to understand whether agents can complete meaningful tasks both correctly and efficiently.}

Consider an agent auditing an expense claim. It must connect the claimed amount in a spreadsheet, the payable amount in a receipt photograph, and the reimbursement rule in a policy document. One agent may inspect the relevant entries, compare the amounts, and submit the correct decision. Another may repeatedly reopen files, pursue unsuccessful queries, and eventually submit the same decision. Their answers receive the same success score, yet the interactions consume different amounts of work and may obtain different evidence. Conversely, additional calls can be justified when a receipt is difficult to read or an apparent duplicate needs verification. Understanding tool efficiency therefore requires examining what interactions accomplish in relation to the task.

This creates two evaluation challenges. First, resource measurements such as call counts and token usage describe interaction volume without identifying which operations recover relevant facts or produce useful artifacts. Second, valid solutions need not follow one reference sequence. A receipt total can be obtained through direct visual inspection, OCR, or a script; a single program can replace several manual operations. Preparation calls may also support later evidence acquisition without directly returning an answer. Evaluation must accommodate these alternative routes while distinguishing recognized progress from repeated, failed, or unrelated work. This motivates the question: How can we assess what tool interactions accomplish while allowing different solution strategies?

We introduce TAE-Bench, a benchmark of 191 multimodal tasks spanning routine execution and strategy reasoning. Its materials include documents, spreadsheets, images, audio, and code, with an outcome checker for each task. We evaluate models through a single fixed harness, rather than each vendor's own agent harness. Shared system instructions and tool interfaces, together with the same initial environment and budgets for each task, establish common operating conditions. This controls variation in the execution framework and tool exposure, making differences in model behavior easier to interpret. Model endpoints, generation settings, and evaluation provenance remain separately documented. Figure~\ref{fig:teaser} contrasts two successful runs on one task and shows how accuracy and effective interaction rate separate across models.

TAE-Bench uses an evidence graph for each task. Nodes represent key facts, intermediate results, and required environment states; edges record their supporting relationships. Explicit predicates check recorded tool observations and state snapshots against these nodes, allowing different tools and sequences to provide evidence for the same task. Reference solutions supply executable examples without defining a unique or optimal route. The graph thus provides an inspectable basis for relating tool interactions to task evidence. Its matches are operational judgments: they do not establish that an agent understood or causally used an observation, and a finite graph can omit useful exploration. Evaluator auditing must therefore examine both useful calls that receive no credit and unrelated calls that receive credit.

We report 27 models on all 191 tasks and organize the analysis around capability, family-associated tool preferences and execution organization. Disclosed scale relates to task success, but does not specify a universal calling style. Models from the same company use more similar tool combinations on jointly solved tasks; directly aligned GPT and Claude comparisons show different material-access and code preferences even though their call totals differ only modestly. Within-company capability gains often accompany fewer calls and less unmatched intermediate work, yet do not consistently increase batching. Generation rounds predict recorded token usage more closely than call totals, linking feedback organization to the resource burden of long and unsuccessful execution. These behavioral observations motivate studying how models map subgoals to tool operations without inferring undisclosed sizes or hidden computation.
Our contributions are:
\begin{itemize}
\item An evaluation formulation that jointly examines task completion, interaction volume, and evidence acquisition while accommodating alternative tool sequences.
\item A multimodal benchmark of 191 tasks, with an evidence graph for each task and a shared harness and tool environment that control variation in the execution framework across models.
\item A 27-model evaluation on all 191 tasks, with matched behavioral comparisons drawn from the full 191-task pool, identifying family-associated tool preferences, version-dependent execution organization, and the relationship between feedback rounds and recorded resource use, together with a public release of the benchmark tasks and execution environment.
\end{itemize}

\begin{figure*}[!th]
  \centering
  \includegraphics[width=\textwidth]{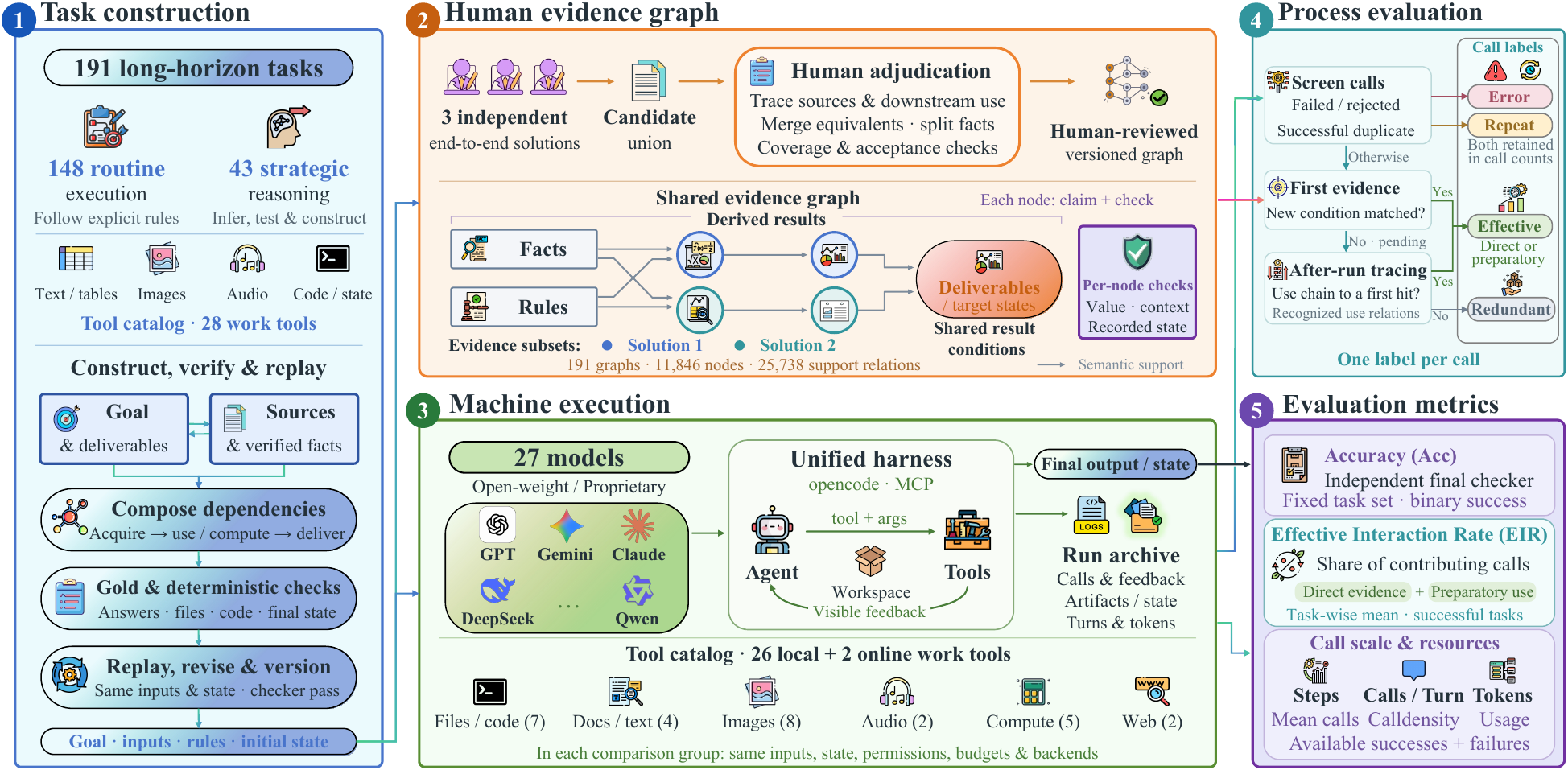}
  \caption{Overview of TAE-Bench. (1)~Tasks are composed, checked
  deterministically, and replayed. (2)~Independent human solutions are merged
  and adjudicated into a versioned evidence graph whose nodes pair a claim
  with a check. (3)~Models run under a unified harness with a shared tool
  catalog, and each run is archived. (4)~Every call receives one label from the
  recorded evidence: \texttt{Effective}, \texttt{Redundant}, \texttt{Repeat},
  or \texttt{Error}. (5)~We report accuracy, effective interaction rate (EIR; panel~5 names its
  complement, the redundancy rate), and call scale and resource use.}
  \label{fig:teb_overview}
\end{figure*}

\section{Related Work}
\label{sec:related_work}
\subsection{Long-Horizon Agent Evaluation}

Agent evaluation has expanded from simple question answering\cite{rajpurkar2016squad,yang2018hotpotqa,hendrycks2021mmlu} to long-horizon tasks, requiring agents to maintain long-term goals and execute reliably over extended interactions.
Early work introduced executable environments to extend evaluation from textual outputs to actual task execution\cite{jimenez2024swebench,drouin2024workarena}.
For example, WebArena\cite{zhou2024webarena} provides a self-hosted web environment in which agents complete tasks through multi-step web interactions, while OSWorld\cite{xie2024osworld} places agents in real operating-system environments to evaluate their computer-use capabilities.
Toolathlon\cite{li2026toolathlon} further increases task complexity through integrating file systems, communication tools, and business applications, requiring agents to reliably transfer information and intermediate results across task stages .
Recent benchmarks treat extended interactions and persistent state management as direct evaluation targets\cite{lu2024toolsandbox,yao2024tau,merrill2026terminalbench,yuan2026osworld2}.
Terminal-Bench 4.0\cite{marten2026terminalbench4} extends evaluation to hundreds of interactions through complex tasks in containerized terminals , while OSWorld 2.0\cite{yuan2026osworld2} introduces dynamic messages, implicit states, and user interactions into hour-scale desktop tasks, challenging agents' ability to sustain task states and adapt to dynamic environments.

\subsection{Agent Efficiency Methods and Metrics}

The execution efficiency of long-horizon tasks has become an important objective in the evolution of model capabilities.
To optimize this, MiMo-V2.6\cite{mimo2026technicalreport} penalizes overly long successful trajectories during agentic reinforcement learning to encourage more concise execution, while GLM-5.3\cite{zai2026glm53} combines post-training with context compaction to reduce output-token consumption per task.
Existing efficiency evaluation of LLM agents mainly relies on aggregate metrics, such as latency\cite{abhyankar2025osworldhuman}, total execution steps\cite{abhyankar2025osworldhuman,li2026toolathlon} and token cost\cite{li2026toolathlon}.
Although these metrics effectively characterize the interaction scale and resource consumption of entire trajectories, they cannot provide fine-grained information about each step's contribution to the task.
To address this limitation, we use evidence-graph-based evaluation in TAE-Bench to quantify each step's contribution to the task, thereby enabling a fine-grained assessment of efficiency.

\section{TAE-Bench}
\label{sec:method}
Beyond task success, TAE-Bench evaluates how efficiently agents use tools and execute tasks (Fig.~\ref{fig:teb_overview}). To this end, we fix the execution environment and tool surface across all
models (Sec.~\ref{sec:tool_environment}); construct 191 long-horizon
tasks whose tool outputs serve verifiable intermediate purposes
(Sec.~\ref{sec:task_construction}); and represent all the evidence an
agent may legitimately need as human-constructed evidence graphs
(Sec.~\ref{sec:evidence_graphs}).
On top of these, we attribute recorded tool calls to graph evidence,
crediting both direct acquisition and the preparatory calls that enable
it (Sec.~\ref{sec:interaction_attribution}).
This yields accuracy, effective interaction rate, and resource
consumption metrics on a unified task set and evaluation environment
(Sec.~\ref{sec:evaluation_metrics}).

\subsection{Unified Harness and General-Purpose Tools}
\label{sec:tool_environment}
\noindent\textbf{Unified execution harness.}
We build a unified harness for TAE-Bench to compare how agents select and
compose shared tool capabilities under controlled task conditions,
ensuring that process differences are attributable to model capability.
Specifically, we use a pinned \texttt{opencode} client~\cite{opencode2026}
as the harness, with its native tools disabled, so every file access,
computation, or retrieval performed by an agent must go through our MCP
service and its 27 tools.
Within each comparison group, task initialization, instructions,
materials, initial checkpoints, permissions, and budgets are held
constant, and all interactions are fully recorded.
This design ensures that the harness itself introduces no capability
differences, so observed process differences are attributable to the
model rather than to the evaluation setup.

\noindent\textbf{Reusable tool interfaces.}
We design tools as general-purpose operations over task materials---the
files, data, and scenes provided by each task---leaving evidence
selection, rule application, and workflow composition entirely to the
agent.
Unlike $\tau$-bench~\cite{yao2024tau}, which exposes domain-specific
business actions, or Terminal-Bench~\cite{merrill2026terminalbench},
which provides terminal execution, we integrate general execution with
material processing within a single interface.
We design 27 tools with distinct functions, covering file access,
documents, images, audio, measurement, retrieval, and computation; the
full tool catalog appears in Appendix A.
Connected through typed interfaces over a shared workspace, tools
preserve their outputs to support compositions such as rendering,
cropping, OCR, and numerical analysis.
We ensure that these tools suffice to complete all TAE-Bench tasks.

\noindent\textbf{Observable tool interactions.}
We record model-visible feedback, the status of each call, produced
artifacts, and the associated state.
These records form the basis for the attribution in
Sec.~\ref{sec:interaction_attribution}: only observable and replayable
interactions can be mapped onto the evidence graph.
Interface, backend, and recording details appear in
Appendix A.

\subsection{Task Construction}
\label{sec:task_construction}
\noindent\textbf{Two complementary task types.}
Long-horizon agents must execute explicit procedures reliably and use
tool-derived evidence to form and test solution strategies. We design TAE-Bench
to cover both demands.
\textbf{Routine (148 tasks): sustained execution.}
Agents apply explicit rules or standard procedures across dependent stages.
\textbf{Strategic (43 tasks): evidence-guided strategy formation.}
Agents additionally infer relationships, compare explanations, or construct
and validate constrained solutions.
All 191 tasks require maintaining information and state across stages.
Tasks span finance, science, engineering, software, transport, media, and
forensics.

\noindent\textbf{Grounded human design.}
These demands guide the human design of materials, rules, result
dependencies, and deliverables. We add evidence graphs to 85 existing tasks
and manually author 106 new tasks from scratch.
For new tasks, authors work backward from verifiable deliverables,
organizing subgoals as
$\mathcal W_\tau=(\mathcal U_\tau,\prec)$, where $u_i\prec u_j$ means that
$u_j$ consumes the result of $u_i$.
Agents discover evidence in materials or environment state through tools
and use it in subsequent execution or judgments.
These dependencies give tool outputs a verifiable task purpose while
permitting alternative tool compositions.

\noindent\textbf{Construction and verification.}
Following environment-and-checker construction
practices~\cite{merrill2026terminalbench,yuan2026osworld2}, humans source
inputs $X$ from public datasets, official records, open media, or controlled
scenes, assemble rules $R$ and initial state $s_0$, and verify reference
facts $Z$ against source annotations, same-source records, original passages,
or construction parameters.
For factual outputs, they derive $y^\star=g_q(Z;R)$.
For all tasks, they specify the complete acceptance set $\mathcal A_\tau$,
implement a deterministic checker, and replay a reference solution from
solver-visible inputs, requiring
\begin{equation}
 \begin{aligned}
 C_\tau(\hat y,s_T)
   &=\mathbf 1\!\left[(\hat y,s_T)\in\mathcal A_\tau\right],\\
 C_\tau(\hat y_{\mathrm{ref}},s_{\mathrm{ref}})&=1.
 \end{aligned}
 \label{eq:task_acceptance}
\end{equation}
Here $(\hat y,s_T)$ denotes the submission and final state;
$\mathcal A_\tau$ covers required answers, artifacts, state, and preservation
constraints.
AI assists organization and implementation; humans design and verify the
tasks.
Independent end-to-end human solutions then supply candidate evidence for
graph construction in Sec.~\ref{sec:evidence_graphs}.
Construction details appear in Appendix B.

\subsection{Human-Constructed Evidence Graphs}
\label{sec:evidence_graphs}
\noindent\textbf{Task-level evidence.}
Final acceptance establishes completion but leaves process differences
unexplained. A single reference call sequence can penalize legitimate tool
substitutions or merged operations.
Building on subgoal-based process analysis~\cite{ma2024agentboard}, we
manually construct evidence graphs of verifiable task claims and their
support relations, providing common evidence conditions for alternative
implementations.

\noindent\textbf{Independent human construction.}
For each of the 191 tasks, $K_{\mathrm h}$ human solvers independently work
through it end to end without consulting one another's initial trajectories.
From their solving processes $H_\tau^{(j)}$, humans pool candidate evidence as
\begin{equation}
 \mathcal P_\tau=\bigcup_{j=1}^{K_{\mathrm h}}\operatorname{Cand}(H_\tau^{(j)}).
 \label{eq:human_candidate_pool}
\end{equation}
They then trace candidates to sources, merge equivalent claims, split compound
facts, and work backward from deliverables to address omissions.
Each admitted node requires a verified source, a concrete task purpose, and
a human-specified acceptance condition.
Humans author nodes and support relations and review each check;
AI assists organization and implementation.
Every graph receives final human review.
The collection contains 11,846 nodes and 25,738 support relations;
Appendix C gives the construction procedure.

\noindent\textbf{Graph representation.}
Each node $v\in V_\tau$ represents a claim $\phi_v$ reviewed by humans:
a fact, rule, derived result, or target state. It has an executable
acceptance check $\chi_v$. Humans identify its supporting parents
$\operatorname{Pa}_\tau(v)$, defining
\begin{equation}
 \begin{aligned}
 G_\tau&=\left(V_\tau,E_\tau,\{\chi_v\}_{v\in V_\tau}\right),\\
 E_\tau&=\{(u,v)\in V_\tau\times V_\tau:
                  u\in\operatorname{Pa}_\tau(v)\}.
 \end{aligned}
 \label{eq:evidence_graph}
\end{equation}
Edges record reviewed support for claims, exposing their grounds and task
purposes. Node checks do not require parents to be matched first.
On available records,
\begin{equation}
 M_{t,v}=\chi_v(a_t,o_t,s_t;h_{t-1})\in\{0,1\},
 \label{eq:node_checks}
\end{equation}
where $o_t$ is model-visible feedback and $s_t$ its associated recorded state.

\noindent\textbf{Multiple valid solutions.}
The replayed reference $H_\tau^{\mathrm{ref}}$ supplies one feasible baseline
that agents may surpass in call count; $G_\tau$ provides result conditions,
while $C_\tau$ independently determines final success~\cite{zhou2024webarena}.
Valid solutions can cover different node subsets through tool substitution
or different derivations. For an aggregate-only deliverable, a direct sum
can replace the reference's group subtotals; full graph coverage is not
required.
Section~\ref{sec:interaction_attribution} uses node checks and observed
result-use relations to attribute first acquisitions and preparatory work
to actual calls.

\subsection{Evidence-Based Process Evaluation}
\label{sec:interaction_attribution}
\noindent\textbf{From evidence to calls.}
On complete records, eligible calls execute successfully, are non-duplicate,
and exclude submission. In recorded order, let $\mathcal I^{(0)}$ contain
eligible calls that first satisfy a previously unmatched human-specified
evidence condition, using $M_{t,v}$ from Sec.~\ref{sec:evidence_graphs}.
Reobtaining a fact through different tools or arguments earns no new direct
credit. Alternative tool choices and derivations are assessed by these
result conditions.

\noindent\textbf{Direct and preparatory contributions.}
After each run, we recognize use relations $D_H$ among eligible calls from
operations and artifact versions, distinct from graph support edges $E_\tau$.
Let $\operatorname{Anc}_{D_H}(S)$ contain eligible calls connected to $S$ by
recognized actual-use chains. We define
\begin{equation}
 \mathcal I^\star=\mathcal I^{(0)}\cup
 \operatorname{Anc}_{D_H}(\mathcal I^{(0)}),\qquad U=|\mathcal I^\star|.
 \label{eq:preparatory_credit}
\end{equation}
Preparatory credit requires a chain reaching $\mathcal I^{(0)}$; creation
alone is insufficient. Each call counts once, even with multiple node
matches. Calls in $\mathcal I^\star$ receive \texttt{Effective}, supplying
$U$ for Sec.~\ref{sec:evaluation_metrics}. The attribution algorithm and an independent
human comparison protocol~\cite{lu2025agentrewardbench} appear in
Appendix D.

\subsection{Evaluation Metrics}
\label{sec:evaluation_metrics}
\noindent\textbf{Accuracy (Acc).}
For fixed tasks $\mathcal T$ and model $m$'s included successes $\mathcal S_m$,
binary $\operatorname{Acc}(m)=|\mathcal S_m|/|\mathcal T|$ gives no partial credit.

\noindent\textbf{Effective Interaction Rate (EIR).}
Call counts alone do not distinguish direct and preparatory contributions
from uncredited work.
With $T_{m\tau}$ non-submission calls and
$U_{m\tau}=|\mathcal I^\star_{m\tau}|$ from
Sec.~\ref{sec:interaction_attribution}, define
\begin{equation}
 \begin{aligned}
 \eta_{m\tau}&=\frac{U_{m\tau}}{T_{m\tau}},\\
 \operatorname{EIR}(m)&=\frac{1}{|\mathcal S_m^{\mathrm{EIR}}|}
 \sum_{\tau\in\mathcal S_m^{\mathrm{EIR}}}\eta_{m\tau}.
 \end{aligned}
 \label{eq:eir}
\end{equation}
Successful, fully labeled runs with $T_{m\tau}>0$ form
$\mathcal S_m^{\mathrm{EIR}}$, conditioning call composition on full
completion. EIR averages per-task proportions equally; the remaining
share, $1-\operatorname{EIR}(m)$, combines \texttt{Redundant},
\texttt{Repeat}, and \texttt{Error} calls.
It counts calls, not nodes, without reference-length normalization.
Full EIR need not imply minimal calls.
Paired EIR comparisons across models $\mathcal M$ use
$\mathcal P_{\mathcal M}^{\mathrm{EIR}}
=\bigcap_{m\in\mathcal M}\mathcal S_m^{\mathrm{EIR}}$.

\noindent\textbf{Tool Calls (Steps).}
Let $\mathcal C_m^g$ contain included successes and failures with metric
$g$'s required records. Absolute call scale is
\begin{equation}
 \operatorname{Steps}(m)=
 \frac{\sum_{\tau\in\mathcal C_m^{\mathrm{step}}}T_{m\tau}}
 {|\mathcal C_m^{\mathrm{step}}|}.
 \label{eq:mean_steps}
\end{equation}

\noindent\textbf{C/T.}
With $J$ counting generations (\texttt{step-finish} events),
\begin{equation}
 \mathrm{C/T}(m)=
 \frac{\sum_{\tau\in\mathcal C_m^{\mathrm{turn}}}T_{m\tau}}
 {\sum_{\tau\in\mathcal C_m^{\mathrm{turn}}}J_{m\tau}}.
 \label{eq:calls_per_turn}
\end{equation}
This measures call density, not concurrency.

\noindent\textbf{Token Consumption (Tokens).}
With confirmed source accounting, $Q_{m\tau}$ sums input, cache-read,
cache-write, output, and reasoning tokens~\cite{kapoor2024agentsmatter}:
\begin{equation}
 \begin{aligned}
 \operatorname{Tokens}_{\mathrm{total}}(m)
 &=\sum_{\tau\in\mathcal C_m^{\mathrm{tok}}}Q_{m\tau},\\
 \overline{\operatorname{Tokens}}(m)
 &=\frac{\operatorname{Tokens}_{\mathrm{total}}(m)}{|\mathcal C_m^{\mathrm{tok}}|}.
 \end{aligned}
 \label{eq:token_aggregation}
\end{equation}
Other paired metrics likewise require common successes and available
records;

\begingroup
\definecolor{tebAccLow}{HTML}{FFF8F7}
\definecolor{tebAccHigh}{HTML}{F29B9E}
\definecolor{tebAccInk}{HTML}{9B4149}
\definecolor{tebEirLow}{HTML}{F5F9FF}
\definecolor{tebEirHigh}{HTML}{98BDF0}
\definecolor{tebEirInk}{HTML}{345F96}
\definecolor{tebGroup}{HTML}{F2F4F5}
\newcommand{\tebacc}[2][]{\cellcolor{tebAccHigh!#2!tebAccLow}#1{#2\%}}
\newcommand{\tebeir}[2][]{\cellcolor{tebEirHigh!#2!tebEirLow}#1{#2\%}}
\newcommand{\teblegend}[2]{\colorbox{#1High!#2!#1Low}{\makebox[1.9em]{\strut #2}}}

\begin{table*}[!t]
\centering
\small
\setlength{\tabcolsep}{5pt}
\renewcommand{\arraystretch}{1}
\caption{Results on all 191 tasks, grouped by official model-family weight releases.
Within each group, models are sorted by strategy-reasoning accuracy in descending order.
Acc. (red) and EIR (blue) use absolute 0--100\% color scales shared across task groups.
EIR averages successful-run values; Steps and Tokens average all runs.
C/T divides total calls by total rounds; K/M denote thousands/millions of tokens.
For Acc., EIR, Steps, and Tokens, \textbf{bold} and \underline{underlining} indicate the best and second-best distinct values in each column across all models, respectively, including ties. C/T is unmarked.}
\begin{tabular}{@{}lccrrr@{\hspace{9pt}}ccrrr@{}}
\toprule
& \multicolumn{5}{c}{\textbf{Long-horizon Strategy Reasoning}}
& \multicolumn{5}{c}{\textbf{Long-horizon Routine Execution}} \\
\cmidrule(lr){2-6}\cmidrule(l){7-11}
Model
& \textcolor{tebAccInk}{Acc.} & \textcolor{tebEirInk}{EIR} & Steps & C/T & Tokens
& \textcolor{tebAccInk}{Acc.} & \textcolor{tebEirInk}{EIR} & Steps & C/T & Tokens \\
\midrule
\rowcolor{tebGroup}
\multicolumn{11}{@{}l}{\strut\textbf{Closed-weight models}} \\
\mlogo{openai}GPT-6 Astra~\cite{openai2026gpt6astra} & \tebacc[\textbf]{65.1} & \tebeir{52.4} & 26.2 & 0.93 & 934K & \tebacc[\textbf]{93.2} & \tebeir[\textbf]{81.8} & \textbf{14.7} & 0.93 & 331K \\
\mlogo{openai}GPT-6.1 Terra~\cite{openai2026gpt61terra} & \tebacc[\underline]{60.5} & \tebeir{50.2} & 24.8 & 1.16 & \textbf{785K} & \tebacc[\textbf]{93.2} & \tebeir{72.1} & 18.6 & 2.07 & \textbf{224K} \\
\mlogo{claude}Claude Opus 5.5~\cite{anthropic2026opus55} & \tebacc[\underline]{60.5} & \tebeir{46.2} & 34.2 & 1.04 & 2.46M & \tebacc{87.8} & \tebeir{75.6} & 17.8 & 1.27 & 743K \\
\mlogo{claude}Claude Opus 5~\cite{anthropic2026opus5} & \tebacc{53.5} & \tebeir{36.9} & 51.6 & 1.03 & 5.08M & \tebacc{87.2} & \tebeir{64.6} & 23.8 & 1.18 & 1.07M \\
\mlogo{openai}GPT-6.1 Sol~\cite{openai2026gpt61sol} & \tebacc{51.2} & \tebeir[\textbf]{60.6} & 23.9 & 0.92 & \underline{813K} & \tebacc[\textbf]{93.2} & \tebeir[\underline]{79.9} & \underline{14.9} & 1.09 & \underline{282K} \\
\mlogo{qwen}Qwen4-Max~\cite{qwen2026qwen4release} & \tebacc{46.5} & \tebeir{44.6} & 57.1 & 1.06 & 6.04M & \tebacc{87.8} & \tebeir{57.2} & 29.4 & 1.40 & 1.15M \\
\mlogo{claude}Claude Sonnet 5.5~\cite{anthropic2026sonnet55} & \tebacc{41.9} & \tebeir{54.6} & \underline{22.6} & 1.05 & 1.30M & \tebacc{66.9} & \tebeir{74.1} & 15.7 & 1.41 & 481K \\
\mlogo{gemini}Gemini-4-Argon~\cite{google2026gemini4argon} & \tebacc{39.5} & \tebeir[\underline]{54.9} & \textbf{21.2} & 0.93 & 1.27M & \tebacc{83.8} & \tebeir{66.1} & 21.1 & 0.92 & 893K \\
\mlogo{openai}GPT-6 Sol~\cite{openai2026gpt6soluna} & \tebacc{34.9} & \tebeir{53.5} & 28.4 & 1.05 & 954K & \tebacc{75.0} & \tebeir{72.2} & 19.9 & 1.72 & 296K \\
\mlogo{claude}Claude Haiku 5.5~\cite{anthropic2026haiku55} & \tebacc{25.6} & \tebeir{41.2} & 36.9 & 1.01 & 3.21M & \tebacc{77.7} & \tebeir{63.2} & 23.7 & 1.41 & 1.09M \\
\mlogo{gemini}Gemini-3.8-Flash~\cite{google2026gemini38flash} & \tebacc{23.3} & \tebeir{28.7} & 85.7 & 0.98 & 4.86M & \tebacc{68.9} & \tebeir{40.0} & 48.9 & 0.96 & 2.09M \\
\mlogo{openai}GPT-6 Luna~\cite{openai2026gpt6soluna} & \tebacc{20.9} & \tebeir{42.2} & 39.6 & 1.13 & 1.89M & \tebacc{53.4} & \tebeir{61.3} & 24.6 & 1.92 & 380K \\
\mlogo{claude}Claude Sonnet 5~\cite{anthropic2026sonnet5} & \tebacc{20.9} & \tebeir{37.1} & 57.0 & 1.08 & 4.91M & \tebacc{64.9} & \tebeir{61.2} & 37.4 & 1.49 & 1.67M \\
\mlogo{gemini}Gemini3.1 Pro~\cite{google2026gemini31pro} & \tebacc{16.3} & \tebeir{37.8} & 74.4 & 0.97 & 4.19M & \tebacc{50.0} & \tebeir{50.4} & 40.0 & 0.95 & 1.54M \\
\mlogo{doubao}Doubao-Seed-2.1-Lite~\cite{volcengine2026modellist} & \tebacc{11.6} & \tebeir{39.1} & 72.6 & 1.01 & 6.10M & \tebacc{51.4} & \tebeir{60.6} & 34.0 & 1.18 & 1.50M \\
\mlogo{doubao}Doubao-Seed-2.1-Pro~\cite{volcengine2026modellist} & \tebacc{7.0} & \tebeir{33.0} & 65.8 & 0.97 & 4.93M & \tebacc{48.0} & \tebeir{56.9} & 33.2 & 0.96 & 1.62M \\
\midrule
\rowcolor{tebGroup}
\multicolumn{11}{@{}l}{\strut\textbf{Open-weight model families}} \\
\mlogo{kimi}Kimi-K3.1-Pro~\cite{kimi2026procard} & \tebacc[\textbf]{65.1} & \tebeir{33.9} & 62.4 & 1.22 & 4.85M & \tebacc[\underline]{90.5} & \tebeir{50.7} & 32.9 & 2.04 & 827K \\
\mlogo{deepseek}DeepSeek-V4.1-Flash~\cite{deepseek2026v41flash} & \tebacc{41.9} & \tebeir{22.3} & 92.7 & 1.06 & 12.58M & \tebacc{81.1} & \tebeir{37.1} & 50.8 & 1.30 & 2.86M \\
\mlogo{qwen}Qwen3.8-Max~\cite{qwen2026maxcard} & \tebacc{41.9} & \tebeir{36.1} & 64.3 & 1.07 & 10.57M & \tebacc{79.7} & \tebeir{53.8} & 30.5 & 1.26 & 1.51M \\
\mlogo{hunyuan}Hunyuan hy4~\cite{tencent2026hy4release} & \tebacc{39.5} & \tebeir{30.6} & 59.8 & 1.03 & 7.90M & \tebacc{80.4} & \tebeir{44.5} & 40.8 & 1.26 & 2.37M \\
\mlogo{kimi}Kimi-K3~\cite{kimi2026k3card} & \tebacc{34.9} & \tebeir{39.6} & 95.8 & 1.06 & 16.50M & \tebacc{78.4} & \tebeir{54.3} & 36.6 & 1.36 & 1.55M \\
\mlogo{qwen}Qwen3.8-Flash~\cite{qwen2026flashcard} & \tebacc{32.6} & \tebeir{32.7} & 92.4 & 1.03 & 18.19M & \tebacc{81.8} & \tebeir{51.9} & 36.5 & 1.15 & 2.13M \\
\mlogo{minimax}MiniMax-M3~\cite{minimax2026m3card} & \tebacc{32.6} & \tebeir{23.4} & 105.6 & 1.08 & 21.39M & \tebacc{66.2} & \tebeir{38.8} & 49.3 & 1.25 & 3.43M \\
\mlogo{glm}GLM-5.3-Flash~\cite{zai2026glm53flashcard} & \tebacc{30.2} & \tebeir{34.8} & 75.7 & 1.07 & 13.58M & \tebacc{76.4} & \tebeir{52.9} & 37.1 & 1.32 & 1.98M \\
\mlogo{mimo}MiMo-V2.6-Pro~\cite{mimo2026technicalreport} & \tebacc{25.6} & \tebeir{39.4} & 67.5 & 1.12 & 10.79M & \tebacc{83.8} & \tebeir{49.7} & 38.0 & 1.40 & 2.44M \\
\mlogo{minimax}MiniMax-M3.1-Flash~\cite{minimax2026m31flashcard} & \tebacc{20.9} & \tebeir{28.2} & 85.3 & 1.06 & 12.93M & \tebacc{65.5} & \tebeir{42.4} & 46.6 & 1.29 & 3.01M \\
\mlogo{mimo}MiMo-V2.6-Flash~\cite{mimo2026technicalreport} & \tebacc{18.6} & \tebeir{39.4} & 49.1 & 1.17 & 6.66M & \tebacc{66.2} & \tebeir{47.0} & 39.6 & 1.46 & 1.95M \\
\bottomrule
\end{tabular}

\vspace{4pt}
{\scriptsize
Acc. (\%):\enspace
\teblegend{tebAcc}{0}\teblegend{tebAcc}{20}\teblegend{tebAcc}{40}\teblegend{tebAcc}{60}\teblegend{tebAcc}{80}\teblegend{tebAcc}{100}
\qquad EIR (\%):\enspace
\teblegend{tebEir}{0}\teblegend{tebEir}{20}\teblegend{tebEir}{40}\teblegend{tebEir}{60}\teblegend{tebEir}{80}\teblegend{tebEir}{100}
\quad Darker = higher.\par}
\label{tab:teb191}
\end{table*}
\endgroup

\section{Experiments and Analysis}
\label{sec:experiments}
\subsection{Experimental scope}
We evaluate 27 models on all 191 tasks under the same base harness and tool environment, yielding 5,157 selected runs (Table~\ref{tab:teb191}).
Metric definitions are in Appendix E; model-wide statistics, further case studies, and release provenance are in Appendices F--H.

The groups differ in the cognitive operations required for their core deliverables. Routine execution (148 tasks) follows supplied rules, standard methods, conventional measurements, or file processing, such as applying a reimbursement policy to receipts. Strategy reasoning (43 tasks) requires at least one core deliverable to infer an unstated structure or cause, assess competing explanations, or construct and validate a solution, such as identifying the holiday calendar behind a spreadsheet's recorded hours.

We count tool calls, including failed attempts, but exclude submit answer. For each task group, C/T divides total calls by total generation rounds across selected runs, including rounds without calls; it is not a mean of per-task ratios. EIR averages available fractions of calls credited by the evidence graph over successful runs. Unless stated otherwise, resource averages include failed runs. Matched comparisons use tasks solved by both models and report the subset size.

\subsection{Overall Results and Model Trends}
\label{sec:overall_results}
Accuracy, evidence attribution, and resource use describe different aspects of execution: high EIR does not imply few calls, and high C/T does not establish tool concurrency.

\begin{table*}[!t]
\centering\small
\setlength{\tabcolsep}{3pt}
\caption{Resource comparisons on identical jointly solved tasks drawn from the full 191-task pool. Shared is the number of jointly successful tasks, which differs across pairs. Calls are means on that subset; differences are M minus C; counterparts are same-family models. EIR differences are in percentage points. Call intervals use paired task bootstrapping.}
\label{tab:paired}
\begin{tabular}{llrrrrl}\toprule
Model (M) & Counterpart (C) & Shared & Calls M & Calls C & $\Delta$ EIR (pp) & $\Delta$ Calls [95\% interval]\\\midrule
\mlogo{openai}GPT-6 Astra~\cite{openai2026gpt6astra} & \mlogo{openai}GPT-6.1 Terra~\cite{openai2026gpt61terra} & 158 & 15.16 & 18.92 & $+8.15$ & $-3.77\;[-4.78,-2.83]$\\
\mlogo{mimo}MiMo-V2.6-Pro~\cite{mimo2026technicalreport} & \mlogo{mimo}MiMo-V2.6-Flash~\cite{mimo2026technicalreport} & 102 & 28.53 & 33.13 & $+4.92$ & $-4.60\;[-6.78,-2.71]$\\
\mlogo{kimi}Kimi-K3.1-Pro~\cite{kimi2026procard} & \mlogo{kimi}Kimi-K3~\cite{kimi2026k3card} & 127 & 26.96 & 29.73 & $-2.56$ & $-2.77\;[-5.32,-0.04]$\\
\mlogo{claude}Claude Sonnet 5.5~\cite{anthropic2026sonnet55} & \mlogo{claude}Claude Haiku 5.5~\cite{anthropic2026haiku55} & 95 & 14.43 & 21.63 & $+10.68$ & $-7.20\;[-9.00,-5.49]$\\
\mlogo{qwen}Qwen4-Max~\cite{qwen2026qwen4release} & \mlogo{qwen}Qwen3.8-Max~\cite{qwen2026maxcard} & 127 & 26.57 & 26.83 & $+4.02$ & $-0.26\;[-2.37,2.09]$\\
\mlogo{gemini}Gemini-4-Argon~\cite{google2026gemini4argon} & \mlogo{gemini}Gemini-3.8-Flash~\cite{google2026gemini38flash} & 105 & 16.43 & 35.95 & $+28.91$ & $-19.52\;[-22.30,-16.89]$\\
\bottomrule
\end{tabular}
\end{table*}

All 27 models have lower accuracy and EIR on strategy tasks, with higher mean calls and recorded tokens. The task groups also reveal different competence profiles. GPT-6 Astra, GPT-6.1 Terra, and GPT-6.1 Sol each solve 138/148 routine tasks, but solve 28, 26, and 22 of the 43 strategy tasks. Claude Haiku 5.5 exceeds Sonnet 5.5 on routine accuracy (77.7\% versus 66.9\%), while Sonnet 5.5 leads on strategy accuracy (41.9\% versus 25.6\%). Aggregate accuracy therefore depends on the task mix.

Models with the same success count can organize external work differently. On routine tasks, Astra and GPT-6.1 Sol average 14.7 and 14.9 calls, with EIRs of 81.8\% and 79.9\%. Terra averages 18.6 calls but places more within each generation (C/T 2.07 versus 0.93 and 1.09), while recording fewer tokens (224K versus 331K and 282K). On strategy tasks, Astra and Kimi-K3.1-Pro each solve 28/43 tasks, but average 26.2 and 62.4 calls, with EIRs of 52.4\% and 33.9\%. These aggregate comparisons do not require the models to solve the same tasks.

Within-family improvements do not follow one resource pattern. Claude Opus 5.5 is more accurate than Sonnet 5.5, while Sonnet uses fewer calls and tokens in both groups. Gemini-4-Argon improves accuracy and EIR over the earlier Gemini endpoints while reducing calls and tokens; its C/T remains near one. Qwen4-Max improves accuracy and EIR over Qwen3.8-Max while reducing calls and tokens in both groups. Kimi-K3.1-Pro improves success and reduces resources relative to Kimi-K3 despite lower EIR. MiMo-V2.6-Pro improves accuracy over Flash but uses more calls and tokens on strategy tasks.

Tool preferences also show a family-associated pattern. For each model pair, we average Jensen--Shannon distances between normalized tool-call distributions on jointly solved tasks. The 30 pairs from the same company average 0.379, compared with 0.446 for 321 pairs from different companies; lower distance indicates a more similar tool mix. Pairs can cover different tasks, and shared providers or adapters may contribute to the association.

\begin{figure*}[t]
\centering
\includegraphics[width=\textwidth]{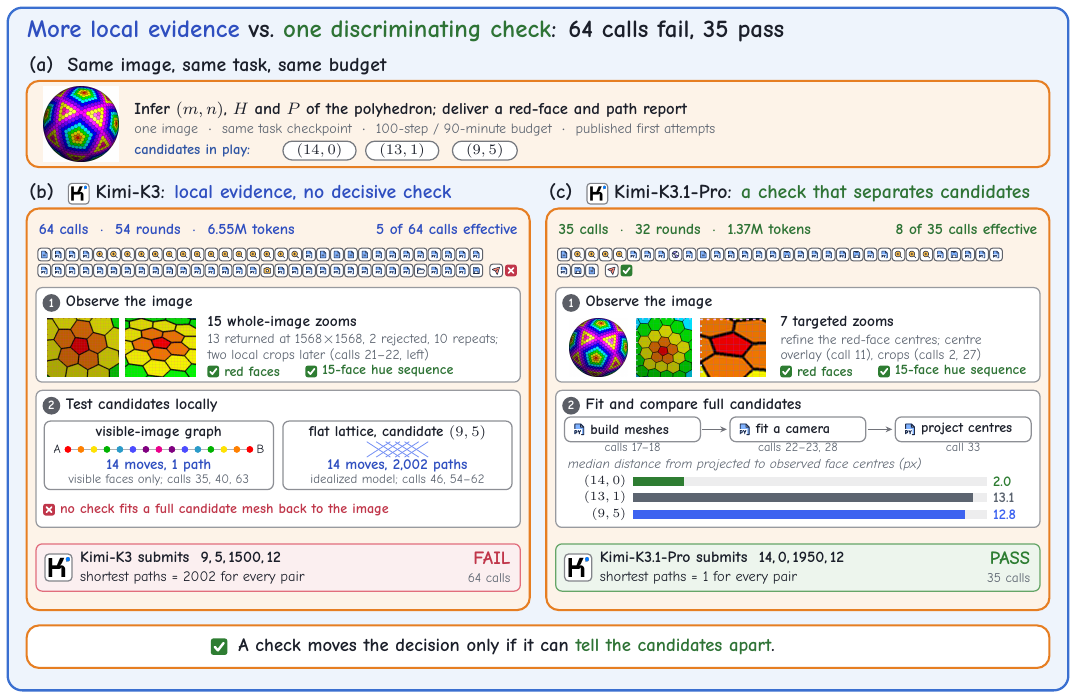}
\caption{\textbf{Organizing tool use around a discriminating test.} Kimi-K3 analyzes local observations and lattice candidates but fails after 64 calls. Kimi-K3.1-Pro compares complete candidate meshes with the image and succeeds after 35 calls. The graph attributes 5 and 8 calls, respectively, to effective evidence acquisition.}
\label{fig:case-polyhedron}
\end{figure*}

\subsection{Efficiency on Shared Successful Tasks}
\label{sec:shared_success}
Conditioning on shared success compares how models complete identical tasks. In Table~\ref{tab:paired}, the model uses fewer calls than its counterpart in all six selected pairs, and five paired call-difference intervals lie below zero. The pairs are selected and their task sets overlap, so these comparisons are descriptive rather than independent tests over all models.

Several pairs combine fewer calls with a higher effective interaction rate. On 158 shared successes, GPT-6 Astra averages 15.16 calls versus 18.92 for GPT-6.1 Terra, with an EIR difference of $+8.15$ percentage points. Claude Sonnet 5.5 averages 14.43 calls versus 21.63 for Haiku 5.5 on 95 shared successes, with an EIR difference of $+10.68$ points. Gemini-4-Argon uses 16.43 calls versus 35.95 for Gemini-3.8-Flash on 105 shared successes, with a $+28.91$-point EIR difference under their evaluated adapter settings.

Call volume and EIR do not always change together. Kimi-K3.1-Pro uses 26.96 calls versus Kimi-K3's 29.73 on 127 shared successes, while its EIR point estimate is 2.56 percentage points lower. Qwen4-Max and Qwen3.8-Max average 26.57 and 26.83 calls on 127 shared successes; their call-difference interval includes zero, while Qwen4-Max has an EIR point estimate 4.02 points higher. These cases show why calls and evidence-credited interaction should be reported together.

These results describe successful executions, while leaving failure behavior and the causes of the differences open. The evidence graph may also leave useful exploration uncredited.

\subsection{Task Understanding and Action Organization}
\label{sec:case_understanding}
Observation scope and call timing offer two ways to organize external evidence. Grouping materials can reduce observations, while batching independent calls can reduce feedback rounds without reducing calls~\cite{lin2026wide,xu2025tpsbench}. We hypothesize that wider batches partly offset the round overhead of narrower observations, but cross-model comparisons cannot establish this.

In a 32-receipt audit, all four models succeed. GPT-6.1 Sol combines the receipts into four sheets, using 14 calls in 12 rounds; GPT-6.1 Terra issues 32 OCR calls in two rounds, using 50 calls in 9 rounds overall. Kimi-K3.1-Pro and DeepSeek-V4.1-Flash issue the OCR calls in four and eight rounds, with 73 calls in 17 rounds and 85 calls in 58 rounds overall, respectively. Calls issued in the same round need not execute concurrently.

DeepSeek already issues four OCR calls per round. Replacing its eight consecutive OCR rounds with two rounds of 16, while holding the rest fixed, would reduce 58 rounds to 52 without reducing calls. This arithmetic assumes independent calls and cannot predict the model's response to different feedback. Most of the round gap to Terra remains. Testing the hypothesis requires varying material grouping and the per-round call limit for the same model and tasks, while measuring correctness, evidence attribution, calls, rounds, and tokens.

Task representation can also shape the actions an agent chooses.
Figure~\ref{fig:case-polyhedron} compares Kimi-K3 and Kimi-K3.1-Pro on the same image-based geometry task. K3's graph of visible faces contains a 14-step path with one shortest route. A separate lattice test predicts 2,002 shortest paths for the $(9,5)$ candidate; K3 nevertheless submits that candidate with 2,002 paths and fails after 64 calls. Its selected candidate conflicts with its own local evidence.

K3.1 Pro constructs candidate meshes, fits a camera view, and projects face centers back onto the image. Its reported median distances to observed centers for $(14,0)$, $(13,1)$, and $(9,5)$ are 2.0, 13.1, and 12.8 pixels, respectively. It submits the correct answer after 35 calls. The evidence graph credits 8 of those calls as effective, compared with 5 of K3's 64 calls.

K3.1 Pro organizes tool use around a global hypothesis test, illustrating how a useful task representation can guide efficient action. These selected runs do not establish a general causal mechanism or an optimal route.

\section{Conclusion}
TAE-Bench connects task success with evidence acquisition under a shared harness. Disclosed scale relates to competence, while model families retain characteristic tool preferences. Stronger variants often compress unmatched intermediate work without adopting one batching strategy. Generation rounds predict recorded tokens more closely than call counts, and failures consume disproportionate resources. These observations motivate studying operation scope and feedback organization through controlled interventions. By examining how agents gather evidence and organize tool calls, TAE-Bench provides a basis for analyzing differences that are hidden by task success alone. The tasks and environment are publicly available.

\bibliography{main}

\clearpage
\beginappendix
\suppressfloats[t]   %
\section{General-Purpose Tools and Execution Details}
\label{app:tools}

\subsection{Scope and Interface Conventions}
\label{app:tool_scope}

The catalog groups the tools of Section~\ref*{sec:tool_environment} into the
six families of Figure~\ref*{fig:teb_overview}: files and code, documents and
text, images, audio, computation, and web. The two web tools,
\texttt{web\_fetch} and \texttt{poi\_search}, are online and available only
to tasks that require network access, so that a model cannot directly search
the external web for answers on local tasks and thereby interfere with
evaluation of its own capability. The remaining tools are local and available
to all tasks.
In addition, we design a separate \texttt{submit\_answer} tool to submit the
answer and end a run.

Each tool has a typed argument schema and a description delivered through
MCP. Results contain structured data or formatted text, optionally
accompanied by images and saved artifact paths. Paths refer to the run's
workspace, exposed to the agent as \texttt{/workspace}. Document pages and
text line indices are one-based. For the image operations below, pixel
coordinates have a top-left origin; bounding boxes use
$[\mathrm{left},\mathrm{top},\mathrm{right},\mathrm{bottom}]$, unless an
explicit normalized-coordinate option is selected.

\subsection{Tool Catalog}
\label{app:tool_catalog}
\newcommand{\tebtoolicon}[1]{%
  \raisebox{-0.28em}{\gfxbyname[height=1.3em]{tool_icons/#1.pdf}}\hspace{0.3em}%
}

The catalog covers these tools and the submission interface. Each tool has a typed argument schema and a description delivered through MCP. We describe each tool's operation, principal inputs, outputs, and relevant controls.

\begin{table*}[t]
\centering
\small
\begin{tabular}{@{}p{0.19\textwidth}p{0.39\textwidth}p{0.36\textwidth}@{}}
\toprule
Tool / operation & Implementation backend & Provided capability \\
\midrule
\tebtoolicon{ocr}\texttt{ocr} & RapidOCR~\cite{rapidai2021rapidocr} / ONNX Runtime & Text recognition and localization \\
\tebtoolicon{object_detect}\texttt{object\_detect} & Grounding DINO~\cite{liu2024groundingdino} & Text-conditioned object localization \\
\tebtoolicon{chart_extract}\texttt{chart\_extract} & DePlot~\cite{liu2023deplot} & Chart-to-table extraction \\
\tebtoolicon{asr}\texttt{asr} & Whisper large-v3-turbo~\cite{radford2023whisper} & Speech transcription and timestamps \\
\tebtoolicon{rag_search}\texttt{rag\_search} & BGE-M3~\cite{chen2024m3embedding} & Embeddings for local passage retrieval \\
\tebtoolicon{audio_beat}\texttt{audio\_beat} & Beat This!~\cite{foscarin2024beatthis}; librosa~\cite{mcfee2015librosa} & Beat/downbeat tracking and tempo estimation \\
\bottomrule
\end{tabular}
\caption{Backend methods and capabilities for the learned-tool subset of the
catalog in Appendix~\ref{app:tool_catalog}. Full model
revisions and dependency versions are provided in the released configuration.}
\label{tab:tool_backends}
\end{table*}

\subsubsection{Files and Code}

\noindent\textbf{\tebtoolicon{read}\texttt{read}.}
Input: a file or directory \texttt{path}, with optional \texttt{offset} and
\texttt{limit}. Returns numbered text lines, directory entries, or a displayed
image. Text reads default to 2,000 lines and are capped at 50~KB; continuation
information supports pagination. PDF and Office content is accessed through
\texttt{document}.

\noindent\textbf{\tebtoolicon{write}\texttt{write}.}
Input: \texttt{path} and full \texttt{content}. Creates parent directories as
needed and writes or replaces the file. Returns a write confirmation. This
operation supports reports, scripts, tables, and other persistent artifacts.

\noindent\textbf{\tebtoolicon{edit}\texttt{edit}.}
Input: \texttt{path}, \texttt{old\_string}, \texttt{new\_string}, and optional
\texttt{replace\_all}. Replaces a matching text span and returns an edit
confirmation. Ambiguous multiple matches require additional context or an
explicit request to replace all occurrences.

\noindent\textbf{\tebtoolicon{glob}\texttt{glob}.}
Input: a filename \texttt{pattern} and optional search \texttt{path}. Returns
up to 100 matching file paths. It locates materials by name or extension
without requiring the agent to read their contents.

\noindent\textbf{\tebtoolicon{grep}\texttt{grep}.}
Input: a regular-expression \texttt{pattern}, optional \texttt{path}, and
optional filename \texttt{include} filter. Returns up to 100 matching text
lines grouped by file with line numbers. Searching inside PDF or Office files
uses \texttt{document} instead.

\noindent\textbf{\tebtoolicon{shell}\texttt{shell}.}
Input: a Bash \texttt{command}, optional \texttt{workdir}, and
\texttt{timeout\_ms}. Executes inside the task container and returns combined
standard output and error with the exit code. The default timeout is 120~s,
with a maximum of 600~s. Each call starts a fresh shell; persistent changes
are passed through files or other task state.

\noindent\textbf{\tebtoolicon{python}\texttt{python}.}
Input: Python \texttt{code} and optional \texttt{timeout\_ms}. Executes a
fresh Python 3.12 interpreter in the task container. Available libraries
include NumPy, pandas, SciPy, openpyxl, matplotlib, and SymPy. Printed output,
execution status, and created files support downstream operations. Timeouts
and output handling follow \texttt{shell}.

\subsubsection{Documents and Text}

\noindent\textbf{\tebtoolicon{document}\texttt{document}.}
Input: \texttt{path}, an \texttt{op}, and operation-specific page, sheet,
range, or query arguments. Operations are \texttt{inspect}, \texttt{read},
\texttt{search}, \texttt{render\_page}, \texttt{tables},
\texttt{read\_sheet}, and \texttt{lines}. Returns metadata, located passages,
page images, table rows, spreadsheet values and formulas, or text lines.
Supports PDF, word-processing documents, slides, spreadsheets, CSV/TSV,
and plain text. Page rendering supports cropping and selectable resolution.

\noindent\textbf{\tebtoolicon{ocr}\texttt{ocr}.}
Input: image \texttt{path}, optional \texttt{bbox}, confidence threshold
\texttt{min\_score}, \texttt{upscale}, and output \texttt{detail}. Returns
recognized Chinese/English text, reading-order lines or boxes, confidence,
and original-image coordinates. The output records the OCR engine actually
used. Region selection and enlargement allow targeted inspection.

\noindent\textbf{\tebtoolicon{rag_search}\texttt{rag\_search}.}
Input: a local \texttt{corpus} file or directory, \texttt{query}, and optional
\texttt{top\_k}, ranking \texttt{mode}, filename filter, and chunking controls.
Supports BM25, dense BGE-M3 retrieval, or hybrid ranking by reciprocal-rank
fusion. Returns passages with file, page, and character-span references.
The corpus index is built on first use and cached.

\noindent\textbf{\tebtoolicon{text_stats}\texttt{text\_stats}.}
Input: an \texttt{op} and inline \texttt{text} or a workspace file
\texttt{path}, with operation-specific arguments such as \texttt{query},
\texttt{pattern}, or \texttt{top\_k}. Supports exact character, word, and
line counts, substring and regular-expression search with positions,
frequency tables, line extraction, character-level comparison, and suspected
misspellings. Returns counts, matches with positions, or differences.

\subsubsection{Images}

\noindent\textbf{\tebtoolicon{zoom}\texttt{zoom}.}
Input: image \texttt{path}, optional \texttt{bbox}, \texttt{scale},
\texttt{normalized}, resampling method, \texttt{grid}, and
\texttt{out\_path}. Returns a displayed crop or enlarged image and its saved
workspace path. An optional grid labels coordinates in the original image,
supporting subsequent measurement and color sampling.

\noindent\textbf{\tebtoolicon{ruler}\texttt{ruler}.}
Input: image \texttt{path}, \texttt{op}, and \texttt{points}, with optional
edge snapping and scale calibration. Operations are \texttt{distance},
\texttt{angle}, \texttt{area}, and \texttt{edges}. Returns segment or
polyline length, angle, polygon area and perimeter, or intensity-edge
positions. Real-world units require \texttt{units\_per\_px} or reference
points with a known length; optional annotation displays the measurement.

\noindent\textbf{\tebtoolicon{color_picker}\texttt{color\_picker}.}
Input: image \texttt{path}, up to 50 pixel \texttt{points}, optional
neighborhood \texttt{radius}, and mean/median \texttt{stat}. Returns RGB,
HEX, HSV, CIE Lab, alpha, and the nearest CSS color with its color distance.
Neighborhood aggregation also reports color spread.

\noindent\textbf{\tebtoolicon{pixel_stats}\texttt{pixel\_stats}.}
Input: image \texttt{path}, optional \texttt{bbox}, and controls for dominant
colors, target colors and tolerance, or histograms. Returns regional RGB
statistics, brightness, contrast, saturation, and requested color shares or
histograms. Named-color bucket thresholds are included in the output.

\noindent\textbf{\tebtoolicon{object_detect}\texttt{object\_detect}.}
Input: image \texttt{path}, English \texttt{labels}, confidence thresholds,
optional \texttt{bbox}, tiling, and annotation controls. Grounding DINO
returns labeled boxes in original-image coordinates, confidence, and
per-label counts. Tiled detection merges overlapping results. Counts are
model estimates conditioned on labels, thresholds, and visible content.

\noindent\textbf{\tebtoolicon{chart_extract}\texttt{chart\_extract}.}
Input: chart image \texttt{path}, optional \texttt{bbox}, and
\texttt{max\_new\_tokens}. DePlot returns an estimated title, table header,
data rows, and raw linearized output. The agent can check extracted values
against visible axes, labels, or other source material before applying task
rules.

\noindent\textbf{\tebtoolicon{barcode_decode}\texttt{barcode\_decode}.}
Input: image \texttt{path} and optional \texttt{bbox}. Returns decoded
contents, symbology, corner coordinates, and available error-correction
information for QR, Data Matrix, Aztec, PDF417, and common one-dimensional
barcodes. An image containing no detected code returns a zero count.

\noindent\textbf{\tebtoolicon{exif}\texttt{exif}.}
Input: media \texttt{path}, optional requested \texttt{tags}, or
\texttt{all\_tags}. ExifTool returns available capture times, GPS,
camera/exposure information, dimensions, orientation, duration, and other
metadata. Missing fields remain null; GPS coordinates use signed WGS-84
latitude and longitude.

\subsubsection{Audio}

\noindent\textbf{\tebtoolicon{asr}\texttt{asr}.}
Input: audio/video \texttt{path}, optional \texttt{language},
\texttt{start\_s}/\texttt{end\_s}, timestamp granularity, and transcription
or translation \texttt{task}. Whisper returns a transcript and requested
segment or word timestamps. Time windows support selective analysis of long
recordings; translation produces English text.

\noindent\textbf{\tebtoolicon{audio_beat}\texttt{audio\_beat}.}
Input: media \texttt{path}, optional time window, \texttt{engine}, and
\texttt{max\_beats}. Returns tempo in BPM, beat/downbeat timestamps, estimated
beats per bar, and tempo stability. Engines include Beat This!, librosa, or
both; the combined mode reports their agreement.

\subsubsection{Computation}

\noindent\textbf{\tebtoolicon{calculator}\texttt{calculator}.}
Input: numeric \texttt{expression}, significant \texttt{digits}, and
\texttt{angle\_unit}. Returns an exact symbolic result and decimal
approximation. Supports arithmetic, roots, logarithms, trigonometry,
combinatorics, and summary statistics; decimal literals are represented as
exact fractions.

\noindent\textbf{\tebtoolicon{datetime_calc}\texttt{datetime\_calc}.}
Input: \texttt{op} and operation-specific dates, time zones, offsets, or
holiday-calendar arguments. Operations are \texttt{convert}, \texttt{diff},
\texttt{add}, \texttt{info}, \texttt{business\_days}, \texttt{holidays},
\texttt{parse}, and \texttt{now}. Returns normalized timestamps, elapsed or
calendar differences, shifted dates, date metadata, or holiday lists.
Time-zone conversion handles daylight saving; \texttt{now} respects a frozen
benchmark clock when configured.

\noindent\textbf{\tebtoolicon{unit_convert}\texttt{unit\_convert}.}
Input: \texttt{value}, \texttt{from\_unit}, and \texttt{to\_unit}. Uses Pint
to return conversions across physical quantities, including offset
temperatures, compound units, data sizes, and supported Chinese market
units. Unit expressions follow Pint syntax.

\noindent\textbf{\tebtoolicon{currency_convert}\texttt{currency\_convert}.}
Input: \texttt{amount}, source/target currency codes, and optional
\texttt{on\_date}. Uses a frozen historical European Central Bank reference-rate
snapshot. Returns the conversion and the latest available fixing date on or
before the requested date; omitting the date selects the snapshot's latest
entry.

\noindent\textbf{\tebtoolicon{hash_codec}\texttt{hash\_codec}.}
Input: an \texttt{op}, inline \texttt{text} or file \texttt{path}, and optional
\texttt{encoding}, \texttt{url\_safe}, or \texttt{save\_to}. Supports MD5,
SHA-1/256/512, SHA3-256, CRC32, Base64/Base32, hexadecimal, URL and Unicode
escaping, and ROT13. Returns a digest or transformed content and an output
file when requested.

\subsubsection{Web}

\noindent\textbf{\tebtoolicon{web_fetch}\texttt{web\_fetch}.}
Input: \texttt{url}, optional \texttt{start\_char}, \texttt{max\_chars},
\texttt{find}, and \texttt{include\_links}. Returns extracted webpage text
as Markdown, either paginated or restricted to snippets around a phrase.
PDFs and images are saved into the workspace and returned by path for
subsequent document or image operations.

\noindent\textbf{\tebtoolicon{poi_search}\texttt{poi\_search}.}
Input: \texttt{op}, keywords, city/location, radius, POI identifier,
coordinate system, and provider as applicable. Operations are
\texttt{input\_tips}, \texttt{text\_search}, \texttt{around\_search},
\texttt{detail}, \texttt{reverse\_geocode}, and \texttt{convert\_coords}.
Returns suggestions, location records, addresses, or transformed coordinates.
Supports AMap and OpenStreetMap-based providers; input locations are
longitude--latitude strings. Coordinate conversion itself requires no
network request, but the interface follows online-tool permissions.

\subsubsection{Submission}

\noindent\textbf{\tebtoolicon{submit_answer}\texttt{submit\_answer}.}
Input: an \texttt{answer} dictionary matching the task's required fields and
types. Validates the submission schema, persists the answer, and closes the
run after a valid submission. A malformed submission returns format
problems for correction. Required output files must already exist when the
agent submits. Correctness is determined afterward by the task's final
checker. Submission is excluded from work-tool call counts.

\subsection{Backend Configuration and Reproducibility}
\label{app:tool_backends}
In our tool setup, tools such as \texttt{ocr} and \texttt{object\_detect}
are driven by specific local models. Table~\ref{tab:tool_backends} lists
the implementation backend model and capability used by each of these tools.

\noindent\textbf{Network response handling.}
The network wrapper supports \texttt{record}, \texttt{replay}, and
\texttt{live} modes. Record mode first checks the response cache, then
fetches and stores eligible uncached responses. Replay mode serves recorded
requests and returns \texttt{REPLAY\_MISS} on a cache miss. Cache keys identify
the tool, provider, and request. A reproduction therefore specifies both the
task's network permissions and the response-handling mode.

\noindent\textbf{Execution and observation records.}
Each run associates a task container and workspace with its allowed tools,
answer schema, and budgets. Tool feedback is returned through the same
protocol layer and recorded alongside arguments, status, and artifact
references. Long generic outputs are capped at 2,000 lines or 50~KB, with
the full text saved for subsequent inspection; tools with their own
pagination apply their specific limits. Returned images retain a
full-resolution artifact while recording the dimensions sent to the model.
The runner also archives the harness session, final artifacts, configured
state snapshots, and tool-service health information, including source,
package, model, and runtime identities.

\section{Task Construction and Validation Details}
\label{app:task_construction}

\begin{figure*}[t]
  \centering
  \includegraphics[width=\textwidth]{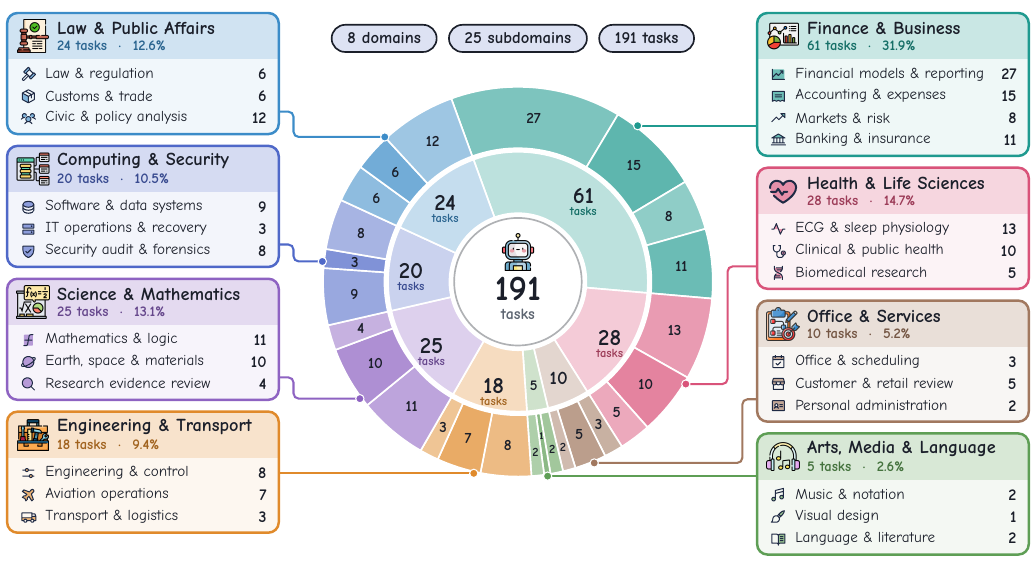}
  \caption{Task domain coverage of TAE-Bench. The inner ring shows the eight
  primary domains and the outer ring their 25 subdomains; sector angle is
  proportional to task count. Linked cards give the subdomain names, counts,
  and each domain's share of the 191 tasks. Each task is counted once, by its
  primary requested judgment or deliverable. The hierarchy is organized by
  the authors from the released domain metadata and task prompts.}
  \label{fig:teb_domain_coverage}
\end{figure*}

\subsection{Task Composition}

TAE-Bench contains a total of 191 tasks, including 106 newly authored tasks;
their construction process is described in Sec.~\ref*{sec:task_construction}
and Appendix~\ref*{app:human_construction_workflow}. In addition, to examine
the adaptability of our evidence graphs, we select 85 tasks from open-source
long-horizon benchmarks for inclusion in TAE-Bench and equip them with the evidence graphs
required for process evaluation.

Following Sec.~\ref*{sec:task_construction}, we divide all tasks into two
categories:

\smallskip\noindent\textbf{Routine (148 tasks): sustained execution.}
All core deliverables can be achieved through explicit rules, unambiguous
standard procedures, computation within a specified model, or ordinary
measurement and file operations, such as applying a reimbursement policy to
receipts.

\smallskip\noindent\textbf{Strategic (43 tasks): evidence-guided strategy formation.}
At least one core deliverable additionally requires inferring an unstated
structure, relationship, or cause; distinguishing competing explanations with
evidence; or constructing and validating a solution under interacting
constraints, such as identifying the holiday calendar behind a spreadsheet's
recorded hours.

Among the 106 newly authored tasks, 96 are routine and 10 are strategic; among
the 85 existing tasks, 52 are routine and 33 are strategic.

\subsection{Human Construction Workflow}
\label{app:human_construction_workflow}

For the 106 newly authored tasks, we jointly define and iteratively refine the
scenario, materials, rules, and acceptance conditions. For the 85 existing
tasks, we verify the inherited materials and requirements before adding
process-evaluation evidence. Independent solving, annotation, and final graph
review cover all 191 tasks.

\noindent\textbf{Design and assemble the task.}
We first specify the user goal and deliverables, including answers, files,
program behavior, or a final environment state. We then work backward from the
deliverables to identify the required observations, operations, and
intermediate results, forming a workflow $\mathcal W_\tau$ that records
dependencies among these results. The workflow does not restrict the choice of
tools, execution order, or intermediate representations.

After defining the task goal, we obtain materials that support the task from
public datasets, official records, openly available media, and software
resources, or construct such materials using known parameters and established
formats. For each answer-related fact, we check whether it can be verified from
the materials; if the materials are insufficient, we adjust the task goal or
supplement or replace the materials. Formally, we denote the materials received
by the solver as $X$ and the verified facts used during construction and
scoring as $Z$. $Z$ may come from original annotations, same-source data
tables, identified source passages, or controlled generation parameters. In
this way, we establish a clear correspondence among the task goal, the
materials visible to the solver, and the basis for scoring.

Finally, we organize the verified materials into task inputs, initialize the
environment, and specify the rules' scope, temporal boundaries, units, and
output format. For all materials, we perform cropping, rendering, re-encoding,
anonymization, and related operations while preserving key content. We also
inspect filenames and auxiliary metadata to avoid leaking the result through
such information.

\noindent\textbf{Build and validate the reference solution.}
After assembling the task, we construct a reference solution and determine the
reference result and acceptance conditions according to the task type. For
factual computation tasks, the reference result is derived independently from
verified facts $Z$ and task rules $R$:
\begin{equation}
  y^\star=g_q(Z;R).
  \label{eq:app_reference_result}
\end{equation}
For file, code, and environment tasks, the reference result includes the
content that the deliverables should contain, the functionality they should
implement, or the final state they should reach. Let $\mathcal A_\tau$ be the
set of submissions and final states that satisfy the complete task
specification, where $\hat y$ denotes the submitted result and $s_T$ denotes
the final environment state. The deterministic checker computes
\begin{equation}
  C_\tau(\hat y,s_T)
  =\mathbf 1\!\left[(\hat y,s_T)\in\mathcal A_\tau\right].
  \label{eq:app_acceptance}
\end{equation}

We then execute the reference implementation from the solver-visible materials
$X$ and initial state, recording its operations and intermediate results to
form a reference execution trajectory. By replaying this trajectory, we check
whether the necessary information can be obtained from the materials, whether
intermediate results support subsequent operations, and whether the final
deliverables are consistent with the reference result and pass the checker. In
this way, we ensure that the task is solvable and the reference trajectory is
executable.

The checker is tested separately using valid equivalent submissions and
targeted incorrect examples: the former should be accepted, and the latter
should be rejected. The checks cover both required changes and preservation
conditions. For example, checks for workbook tasks examine edited cells,
unchanged regions, and retained charts; checks for environment tasks examine
the requested configuration and preserved path ordering. If the reference
execution fails or the check results are inconsistent with the task rules, we
revise the materials, task specification, environment, reference trajectory, or
checker and revalidate the affected parts.

Finally, for each task, we obtain a feasible reference execution trajectory, a
verified reference result, and a set of tested acceptance checks, providing a
foundation for subsequent independent solving and evidence-graph annotation.

\noindent\textbf{Solve each task independently.}
We further assign three human experts ($K_{\mathrm h}=3$) to complete each task
independently from the given materials and rules. Only afterward do they
compare their initial solutions, yielding a total of $191\times3=573$ complete
human solutions. We aggregate the evidence from the three solutions for each
task to cover valid findings beyond those in the reference solution.

\noindent\textbf{Annotate claims and support relations.}
Based on the reference solution and the three independent solutions, we split
candidate claims into nodes that can be independently verified and trace each
node to a specific source, such as a document passage, spreadsheet cell, image
region, audio interval, or recorded environment state. For each node, we check
its object, value, units, and scope and determine how it supports the final
deliverable. We merge equivalent findings and separately retain valid evidence
required by different solutions; missing evidence is identified by working
backward from the deliverables and then added. For each support relation, we
check whether the parent node actually provides the input, rule, or
intermediate result needed to justify the child node.

\noindent\textbf{Review each graph before freezing it.}
Finally, we specify the acceptable forms of evidence for each node and use
valid equivalent forms and targeted incorrect examples to review the
executable checks. For example, when checking an amount, we should not accept
an identical amount appearing on a different receipt. We check evidence
coverage against the three independent solutions and the verified reference
implementation, and correct missing support relations and checks that are too
broad.

We use AI to assist with task construction, but only to organize annotations
and implement checks specified by humans; all 191 evidence graphs receive final
human review. The released evidence graphs contain 11,846 nodes and 25,738
support relations (Appendix~\ref{app:graph_construction}). We record the
versions of the task specification, materials, environment, reference solution,
checker, and evidence graph together so that each run can be matched to the
task assets it actually used.

\subsection{Application and Material Breadth}

Tasks span finance and business, law and public affairs, health and life
sciences, science and mathematics, engineering and transport, computing and
security, office and services, and arts, media, and language. Inputs combine
documents, spreadsheets, images, maps, audio, source code, databases, and file
environments. Deliverables include structured answers and audit lists, edited
workbooks, queries, programs, and environment state.
Figure~\ref{fig:teb_domain_coverage} groups the 191 tasks into eight primary
domains and 25 subdomains.

\section{Evidence Graph Construction and Review}
\label{app:graph_construction}

\subsection{Human Construction Scope}

We set $K_{\mathrm h}=3$: three human experts independently solve each of
the 191 tasks in full, followed by manual graph construction and review.
Independence applies to the initial
solving stage: each person works through the task without consulting the
others' initial trajectories. During adjudication, the team compares the
candidate results with the original materials, rules, and deliverables.
AI assists material organization and check implementation; humans determine
the admitted claims, their task purposes, and their acceptance conditions.

The released graphs contain 11,846 nodes and 25,738 support relations. The checks comprise 3,966 \texttt{answer\_match} nodes and 7,880
\texttt{rule\_check} nodes. These are implementation mechanisms for the
human-authored conditions, covering visible values or expressions and
task-specific output or state checks. Node identifiers are unique within each graph and parents precede their children. Before release, each reference solution is replayed with full checker credit and its required calls are recognized by the graph.

\subsection{From Independent Solutions to Reviewed Graphs}

Procedure~\ref{proc:human_graph} makes the human decisions in graph
construction explicit. Candidate pooling is followed by source and purpose
adjudication, including additions needed to support the deliverables.

\begin{humanprocedure}[tbp]
\caption{Human Construction and Review of Evidence Graphs}
\label{proc:human_graph}
\begin{algorithmic}[1]
\Require Task $\tau$: materials $X$, rules $R$, acceptance requirements
$\mathcal A_\tau$; $K_{\mathrm h}$ experts.
\Ensure Reviewed $G_\tau=(V_\tau,E_\tau,\{\chi_v\}_{v\in V_\tau})$ and version.
\State \textbf{Experts:} Independently solve $\tau$ in full, yielding
$H_\tau^{(1)},\ldots,H_\tau^{(K_{\mathrm h})}$.
\State \textbf{Humans:} Extract candidate claims and pool
$\mathcal P_\tau=\bigcup_{j=1}^{K_{\mathrm h}}\operatorname{Cand}(H_\tau^{(j)})$.
\State \textbf{Humans:} Verify each candidate's source, meaning, and task purpose.
\State \textbf{Humans:} Merge equivalent claims and split compound claims into
independently checkable statements.
\State \textbf{Humans:} Work backward from deliverables to add missing support
and justified alternative branches.
\State \textbf{Humans:} Admit verified nodes $V_\tau$ and annotate supporting
parents $\operatorname{Pa}_\tau(v)$ to define $E_\tau$.
\State \textbf{Humans:} Specify and review each $\chi_v$; AI may assist implementation.
\State \textbf{Humans:} Check coverage and behavior against the independent
solutions and available verified implementations; give final review and record
the version.
\end{algorithmic}
\end{humanprocedure}

\noindent\textbf{Source and support checks.}
Reviewers locate the supporting page, spreadsheet cell, image region, audio
interval, database record, or file version, checking object, value, unit,
time, and context. Negative findings specify the searched scope and
conditions. A candidate found by one solver can be admitted when its source
and purpose are verified; agreement does not replace source checking.
Each edge requires a substantive input, rule, or derivation relationship.
Temporal adjacency or access to the same file is insufficient. The resulting
support annotations preserve alternative branches without concatenating the
human trajectories into a prescribed path.

\noindent\textbf{Acceptance and confirmation.}
Valid equivalent forms and targeted incorrect cases check the intended
conditions: a value from the wrong receipt or an unrelated occurrence of the
same number should not establish the target fact. Humans distinguish a
solution's result from whether its executable check recognizes it, reviewing
revisions with the task version. Trajectories used for graph construction
or revision remain separate from independent trajectory validation used to
assess attribution accuracy.

\subsection{Node and Edge Semantics}
\label{app:graph_semantics}

\noindent\textbf{Executable node records.}
Each serialized node stores \texttt{id}, \texttt{from}, and \texttt{check}.
The notation $\phi_v$ denotes the task claim associated with node $v$;
constructors verify its source and task purpose and specify its acceptance
condition $\chi_v$ through \texttt{check}. A fact or rule can be grounded
in an original material, while a derived result or target state is checked
against the task's inputs, requirements, and recorded outputs or state.
Recognition by $\chi_v$ is an operational check on available records.

\noindent\textbf{Directed support annotations.}
The parents in Eq.~\ref*{eq:evidence_graph} correspond to the serialized
support list:
\begin{equation}
 \operatorname{Pa}_\tau(v)=
 \begin{cases}
 \varnothing,&\texttt{from}(v)=\texttt{null},\\
 \operatorname{set}(\texttt{from}(v)),&\text{otherwise}.
 \end{cases}
 \label{eq:app_support_parents}
\end{equation}
An edge $u\to v$ records a reviewed input, rule, or other basis supporting
$v$. The schema requires parents to precede their children in storage,
providing a topological ordering of the annotation. This ordering does not
prescribe the model's call sequence. Multiple incoming edges expose
supporting grounds; the node checks do not implement a requirement to match
every parent before evaluating the child. These task-level annotations are
distinct from the actual artifact-use relations $D_H$ in
Sec.~\ref*{sec:interaction_attribution}.

\noindent\textbf{A recorded receipt-audit subgraph.}
In \texttt{sroie\_claims\_v1}, let $r$, $p$, and $c$ denote
\texttt{0017\_receipt\_c04}, \texttt{0008\_policy\_alcohol}, and
\texttt{0053\_claim\_c04}, respectively. The recorded parent set is exactly
$\operatorname{Pa}_\tau(c)=\{r,p\}$:
\begin{equation}
 \begin{array}{rcl}
 \phi_r\;\text{(C04 receipt)} & \searrow & \\
 && \phi_c\;\text{(C04: ALCOHOL)}\\
 \phi_p\;\text{(alcohol rule)} & \nearrow &
 \end{array}
 \label{eq:app_receipt_support}
\end{equation}
The receipt evidence is grounded in \texttt{IMG\_5728.jpg}; the policy
provides the applicable alcohol rule. Together they support the recorded
classification. The receipt check accepts either its supported textual
alternative or a view of the relevant original image, so this support
annotation does not prescribe an OCR step. The classification node checks
the joint occurrence of \texttt{C04} and \texttt{ALCOHOL}. This local example
shows how humans organize evidence and its purpose while leaving tool
choice to the solver.

\subsection{Reference Implementation and Alternative Solutions}

The three independent human solutions supply candidate evidence. Separately,
the published $H_\tau^{\mathrm{ref}}$ is one implementation organized,
verified, and replayed by the constructors.
Its frozen work-tool call count provides a feasible comparison anchor. An agent can satisfy the same complete
acceptance conditions with fewer calls through alternative reasoning,
batching, or tool composition.

For example, a receipt-audit graph can contain the claimed amount, the
receipt's payable total, the applicable reimbursement rule, and the approved
amount. Humans distinguish the payable total from cash tendered and change,
check the corresponding fields, and verify how the rule determines the
approved amount. OCR and visual inspection can establish the same receipt
fact. A reference implementation may compute group subtotals, while another
valid implementation directly computes the required aggregate. Relevant
intermediate results are retained as solution-specific branches; a successful
run need not realize every branch.

Three solvers broaden candidate sources, while source, meaning, and purpose
adjudication determine admission. Previously unseen claims or implementations
are reviewed against the same task requirements and can lead to versioned
graph or rule revisions. This procedure supports alternative solutions. Final acceptance remains governed by $C_\tau$.

\subsection{From Semantic Conditions to Call Attribution}

On a run with the required records, let $\mathcal S_H$ index the successful,
nonduplicate, non-submission calls eligible for matching. The node checks
define the detected set
\begin{equation}
 \begin{aligned}
 V_{\mathrm{obs}}(H_\tau)
 &=\{v\in V_\tau:\exists t\in\mathcal S_H,\ M_{t,v}=1\}\\
 &\subseteq V_\tau.
 \end{aligned}
 \label{eq:app_observed_nodes}
\end{equation}
This set describes detected evidence and does not define an intermediate
accuracy score or a full-graph completion threshold. When required outputs,
images, or state records are unavailable, non-verifiability is distinguished
from a failed check. The observed matrix $M$ is an input to contribution
attribution.

Graph edges $E_\tau$ record reviewed semantic support. The actual production
and use of temporary scripts, files, or other preparatory artifacts is
checked from the run's records in Sec.~\ref*{sec:interaction_attribution}.
Such artifacts can contribute without being individually enumerated in the
human graph. Node recognition, first acquisition, repeated acquisition, and
preparatory use remain distinct operations; one call that establishes
multiple nodes still counts as one call.

\section{Contribution Attribution and Validation Protocol}
\label{app:attribution_validation}

\subsection{Observation Sources and Call-Level Rules}

The attribution unit is a model-issued work-tool call, excluding final
submission. Internal reasoning and ordinary assistant messages are not
counted as tool calls. Harness sessions provide the preferred record of
arguments, execution status, and the text and images actually shown to the
model. Service logs and associated snapshots supply execution and state
evidence. Rejected calls that never reach the tool service are retained from
the harness session. Alternative log sources require checks of visibility
and call coverage before being treated as equivalent.

Immediate evidence is evaluated from model-visible feedback. If long
output is truncated and the full text saved to a file, the hidden portion
enters the visible history when it is subsequently read. State-based rules
can check recorded file or environment changes without requiring the full
evaluation-side snapshot to have been shown to the model. Associated
snapshots are checked with their timing and artifact versions; a call index
alone does not establish an isolated causal state transition when operations
overlap.

Let $\mathrm{ok}_t$ indicate execution success and $\kappa_t$ be the canonical
serialization of the tool name and arguments.
The frozen scorer's duplicate-call rule is
\begin{equation}
 r_t=\mathbf 1\!\left[
 \kappa_t\in\{\kappa_j:j<t,\ \mathrm{ok}_j=1\}\right].
 \label{eq:app_duplicate_call}
\end{equation}
Failed calls are not added to the successful-call set, allowing a successful
retry to proceed to node matching. Duplicate detection compares tool and normalized arguments; it is distinct from whether a later observation
contains a previously matched fact. Calls using a different tool or arguments
can still fail to acquire new graph evidence.

Eligible calls form $\mathcal S_H=\{t:\mathrm{ok}_t=1,\ r_t=0\}$.
For these calls, the checks from Sec.~\ref*{sec:evidence_graphs} identify
all newly matched nodes. With $F_{t-1}$ the previously matched nodes,
\begin{equation}
 \begin{aligned}
 N_t&=\{v\in V_\tau\setminus F_{t-1}:M_{t,v}=1\},\\
 F_t&=F_{t-1}\cup N_t,\qquad F_0=\varnothing.
 \end{aligned}
 \label{eq:first_evidence}
\end{equation}
For excluded calls, $N_t=\varnothing$: execution errors and duplicate calls
bypass node matching. The directly contributing set is
$\mathcal I^{(0)}=\{t\in\mathcal S_H:N_t\ne\varnothing\}$.
A call can first match several nodes, and each first acquisition is retained;
the call itself still contributes at most one effective-call count.
Direct evidence can remain valid local progress in an ultimately
unsuccessful run. Final task success is determined separately by $C_\tau$.

\subsection{Recognized Use Relations and Preparatory Credit}

$D_H$ records use relations recognized from the particular run, separately
from the reviewed semantic support edges $E_\tau$. The implementation
examines call arguments and supported Python and shell structures to
identify file production and use, creation and subsequent use of a
directory, and discovery of an initially unspecified file or directory name
followed by its use. Supported file uses include reading, execution,
importing, loading, viewing, and recognized conversions.

File relations are checked against versions. Credit follows the last
identifiable write before use; when that write modifies a file in place,
the protocol can also retain the preceding complete write. Directory and name
discovery have corresponding checks against the prompt, earlier observations,
and initial workspace.

Algorithm~\ref{alg:call_attribution} combines first acquisitions, recognized
preparatory use, and label assignment on complete, evaluable records.
Node tests retain the observation and provenance rules, including image
source, overwrite, and blank-image checks. The helper
\textsc{RecognizeUses} implements only the supported relations described
above, retaining their consumers and artifact versions.
$\operatorname{Pred}_{D_H}$ returns eligible immediate predecessors.

\begin{algorithm}[tbp]
\caption{Evidence-Based Call Attribution}
\label{alg:call_attribution}
\begin{algorithmic}[1]
\Require Frozen $G_\tau$; complete, evaluable non-submission calls
         $H_\tau=(a_t,o_t,s_t)_{t=1}^{T}$.
\Ensure Call labels, effective-call count, and attribution basis.
\State Initialize $S,F,\mathcal S_H,\mathcal I^{(0)}$ as empty sets.
\For{$t=1,\ldots,T$}
  \State $N_t\gets\varnothing$
  \If{$\mathrm{ok}_t=0$}
    \State $\ell_t\gets\mathrm{Error}$
  \ElsIf{$\kappa_t\in S$}
    \State $\ell_t\gets\mathrm{Repeat}$
  \Else
    \State $S\gets S\cup\{\kappa_t\}$; $\mathcal S_H\gets\mathcal S_H\cup\{t\}$
    \State Evaluate $M_{t,v}$ for $v\in V_\tau\setminus F$ under observation and source rules.
    \State $N_t\gets\{v\in V_\tau\setminus F:M_{t,v}=1\}$
    \State $F\gets F\cup N_t$
    \If{$N_t\ne\varnothing$}
      \State $\mathcal I^{(0)}\gets\mathcal I^{(0)}\cup\{t\}$
    \EndIf
  \EndIf
\EndFor
\State $D_H\gets\Call{RecognizeUses}{H_\tau,\mathcal S_H}$
\State $\mathcal I^\star\gets\mathcal I^{(0)}$
\While{$\operatorname{Pred}_{D_H}(\mathcal I^\star)\nsubseteq\mathcal I^\star$}
  \State $\mathcal I^\star\gets\mathcal I^\star\cup\operatorname{Pred}_{D_H}(\mathcal I^\star)$
\EndWhile
\State Label $\mathcal I^\star$ as \texttt{Effective}.
\State Label $\mathcal S_H\setminus\mathcal I^\star$ as \texttt{Redundant}.
\State $U\gets|\mathcal I^\star|$
\State \Return $\{\ell_t\}_{t=1}^{T}$, $U$, $\{N_t\}_{t=1}^{T}$, and $D_H$.
\end{algorithmic}
\end{algorithm}

A recognized chain such as script creation, execution, and subsequent
reading of its output can credit preparatory calls when it reaches task
evidence. Unused artifacts receive no preparatory credit. Errors and
duplicates retain their original labels; each call receives one label.
The non-effective share, $1-\operatorname{EIR}$, combines \texttt{Error},
\texttt{Repeat}, and \texttt{Redundant}.

Missing records and a negative check are distinguished. When required
state is unavailable or attribution fails, diagnostics and independently
available call counts are retained, while the affected attribution metrics
are unavailable under the evaluation protocol. Final correctness remains
separately evaluable when its own required evidence is present.

\subsection{Independent Human Comparison}

This protocol specifies a separate audit of call attribution on model
trajectories selected for confirmation. The three independent human
solutions and final human review of every graph are part of benchmark
construction (Appendices~\ref{app:task_construction}
and~\ref{app:graph_construction}). The audit protocol below sets out how
to measure agreement and report coverage on confirmation trajectories.

\noindent\textbf{Freeze assets and select confirmation runs.}
The audit manifest identifies tasks and materials, graph and check versions,
attribution code and parameters, harness sessions, images, and state records.
Runs used for graph construction, revision, or rule calibration are tracked
as development material. Confirmation runs are selected from records verified
not to have contributed to those adjustments. Selection records describe
coverage across tasks, models, tools, and success states. Representative
sampling and targeted challenge cases are reported separately.

\noindent\textbf{Label complete trajectories independently.}
Two reviewers independently inspect the task materials and full trajectory,
including prior visible history, actual feedback, associated state, and
downstream artifact use. Automatic labels and node matches are withheld
until initial judgments are locked. Reviewers label all non-submission
calls, including errors and repeated attempts, and record the evidence for
each decision. Direct evidence and preparatory use are distinguished.
Insufficient evidence remains undetermined. Disagreements are adjudicated
with recorded reasons; prior involvement in construction or exposure to
labels is documented.

\noindent\textbf{Measure missed contributions and false credit.}
On the paired calls $\mathcal P$ with an adjudicated human judgment and a
valid automatic label, let $b_i=1$ denote human-recognized contribution and
$\hat b_i=1$ denote automatic \texttt{Effective}. The comparison measures
\begin{equation}
 \begin{aligned}
 e_{\mathrm{miss}}
 &=\frac{\sum_{i\in\mathcal P}\mathbf 1[b_i=1\land\hat b_i=0]}
 {\sum_{i\in\mathcal P}\mathbf 1[b_i=1]},\\
 e_{\mathrm{extra}}
 &=\frac{\sum_{i\in\mathcal P}\mathbf 1[b_i=0\land\hat b_i=1]}
 {\sum_{i\in\mathcal P}\mathbf 1[b_i=0]}.
 \end{aligned}
 \label{eq:app_attribution_errors}
\end{equation}
A miss rate $e_{\mathrm{miss}}$ of at most 10\% is treated as acceptable.
Zero denominators are reported as unavailable. Missing automatic labels,
undetermined human judgments, and unaligned records are reported separately.
Reporting includes initial reviewer agreement, adjudication, label confusion,
coverage, and per-task error summaries. Differences in effective-call
proportions are evaluated on the same fully judgeable successful
trajectories with the same aggregation weights.

\noindent\textbf{Check alternative implementations and revise transparently.}
Paired cases examine tool substitutions, equivalent values and formats,
different batching granularity, updated-state rereads, used versus unused
artifacts, bounded exclusion findings, and required verification. They also
test unrelated occurrences of the same value and content absent from the
model-visible output. Unfamiliar algorithms or intermediate representations
are judged against source facts, task requirements, and actual uses.
Problems lead to documented graph or rule revisions and a new frozen
version applied consistently to the comparison set. Confirmation for the
new version uses runs not involved in that revision.

\section{Evaluation Protocol and Metric Definitions}
\label{app:evaluation_protocol}

\subsection{Comparison Sets and Run Inclusion}

Each comparison identifies its fixed planned task set $\mathcal T$, models,
task and scoring versions, run provenance, and record coverage. Let
$\mathcal C_m\subseteq\mathcal T$ denote model $m$'s tasks with a run included
under the stated protocol, and $\mathcal S_m\subseteq\mathcal C_m$ the tasks
formally judged successful. All runs share the execution conditions described in Sec.~\ref*{sec:tool_environment}.

One designated record is included per model--task pair according to the
source's selection rule. Task failure does not trigger selection of a
higher-scoring rerun. Infrastructure and provider failures are handled separately from task failures. Timeouts, context overflow, truncated outputs,
and missing submissions are recorded separately from infrastructure faults.

Task specifications set tool-call and wall-clock budgets. Service-side
budget accounting includes submission; metric call counts exclude final
submission and include model-issued errors and harness-rejected calls.
Full sessions provide the preferred call count. Missing records are not silently treated
as measured zero calls or consumption.

\subsection{Binary Completion and Contribution Composition}

\noindent\textbf{Accuracy.}
The final checker gives binary success, regardless of local progress or
diagnostic partial scores. Accuracy uses the fixed comparison denominator:
\begin{equation}
 \operatorname{Acc}(m)=\frac{|\mathcal S_m|}{|\mathcal T|}.
 \label{eq:app_accuracy}
\end{equation}
Unjudged or absent records are reported separately from judged failures;
they do not change the planned denominator. The checker can verify answers,
files, functionality, environment state, and any explicitly required
operations.

\noindent\textbf{Effective interaction rate.}
For a run on task $\tau$, $T_{m\tau}$ counts all non-submission tool calls
and $U_{m\tau}$ counts calls labeled \texttt{Effective} in a fully labeled
run, including first evidence acquisitions and recognized preparatory
contributions. The main
process-summary set is
\begin{equation}
 \begin{aligned}
 \mathcal S_m^{\mathrm{EIR}}=\{\tau\in\mathcal S_m:
 &\ T_{m\tau}>0,\\
 &\ \text{all }T_{m\tau}\text{ call labels are available}\}.
 \end{aligned}
 \label{eq:app_eir_eligible_tasks}
\end{equation}
Equation~\ref*{eq:eir} first computes the fraction of effective
calls in each eligible run, then averages across tasks with equal weights.
The remaining calls combine \texttt{Redundant}, \texttt{Repeat}, and
\texttt{Error}; successful executions without a recognized task contribution
and failed executions remain separately identifiable in the labels.
An empty summary set yields an unavailable EIR. A confirmed zero-call run
can contribute to accuracy and call-count summaries but has no EIR.

EIR measures contribution composition under the fixed graph and attribution
protocol. Graph nodes and support relations are not counted as calls, and
the reference implementation does not determine $U_{m\tau}$. Preparatory
work can receive credit through its recognized downstream use.
Appendix~\ref{app:attribution_validation} specifies a separate protocol
for assessing attribution reliability.

Failure trajectories can retain correctly acquired local evidence in their
labels; the main EIR summary conditions on successful completion.

\subsection{Call Scale and Generation Turns}

Let $\mathcal C_m^{\mathrm{step}}$ contain included tasks with a confirmed
call count, and $\mathcal C_m^{\mathrm{turn}}$ those with both call and turn
counts. Steps and C/T are
\begin{equation}
 \begin{aligned}
 \operatorname{Steps}(m)
 &=\frac{\sum_{\tau\in\mathcal C_m^{\mathrm{step}}}T_{m\tau}}
 {|\mathcal C_m^{\mathrm{step}}|},\\
 \mathrm{C/T}(m)
 &=\frac{\sum_{\tau\in\mathcal C_m^{\mathrm{turn}}}T_{m\tau}}
 {\sum_{\tau\in\mathcal C_m^{\mathrm{turn}}}J_{m\tau}}.
 \end{aligned}
 \label{eq:app_call_scale}
\end{equation}
Both include successful and failed runs with the required records. In the
current harness, a generation turn corresponds to a recorded
\texttt{step-finish} event. The denominator includes turns with zero tool
calls. C/T describes calls per generation as a ratio of totals.
Execution concurrency requires timing records.

EIR and call scale answer different questions. Two successful implementations
can contain only credited calls and therefore both have full EIR, while
using different numbers of operations. EIR is therefore interpreted
jointly with absolute calls and resource use.

\subsection{Token Components and Aggregation}
\renewcommand{\dbltopfraction}{0.72}
\renewcommand{\dblfloatpagefraction}{0.9}%
\makeatletter\setlength{\@dblfptop}{0pt}\setlength{\@dblfpsep}{14pt plus 2pt}\makeatother

\begin{figure*}[t]\centering
\includegraphics[width=\textwidth]{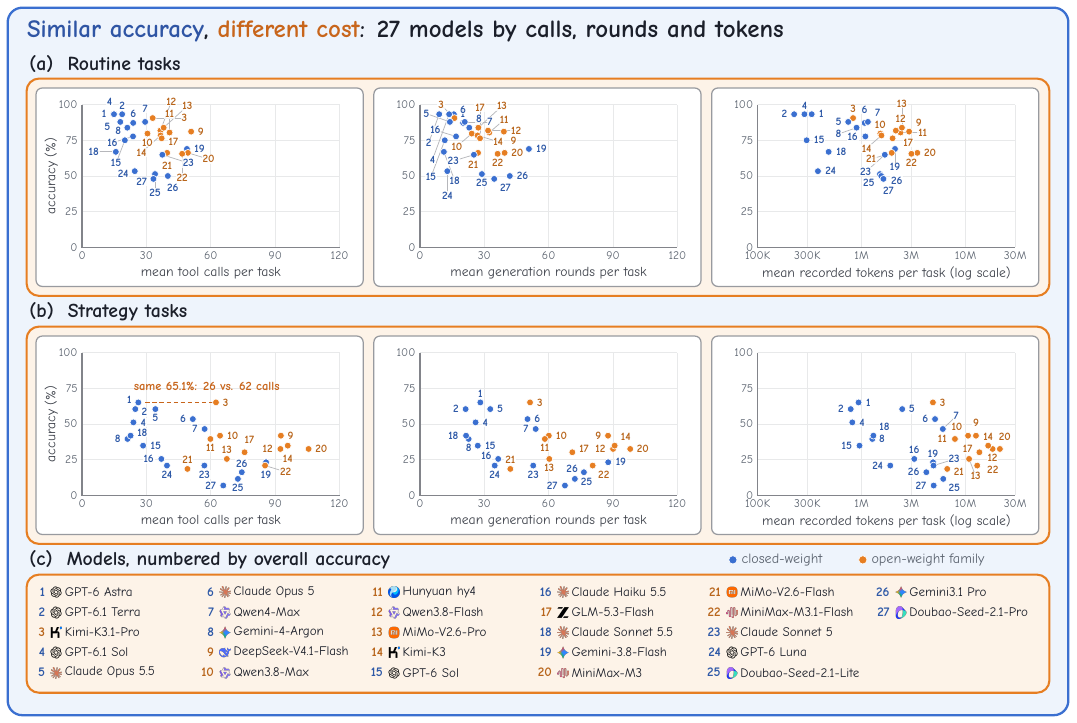}
\caption{\textbf{Similar accuracy, different cost.} Accuracy of the 27 models against mean calls, rounds, and recorded tokens per task, for (a) routine and (b) strategy tasks. Means include failed runs; token accounting differs across providers.}
\label{fig:stat-efficiency}\end{figure*}

\begin{figure*}[t]\centering
\includegraphics[width=\textwidth]{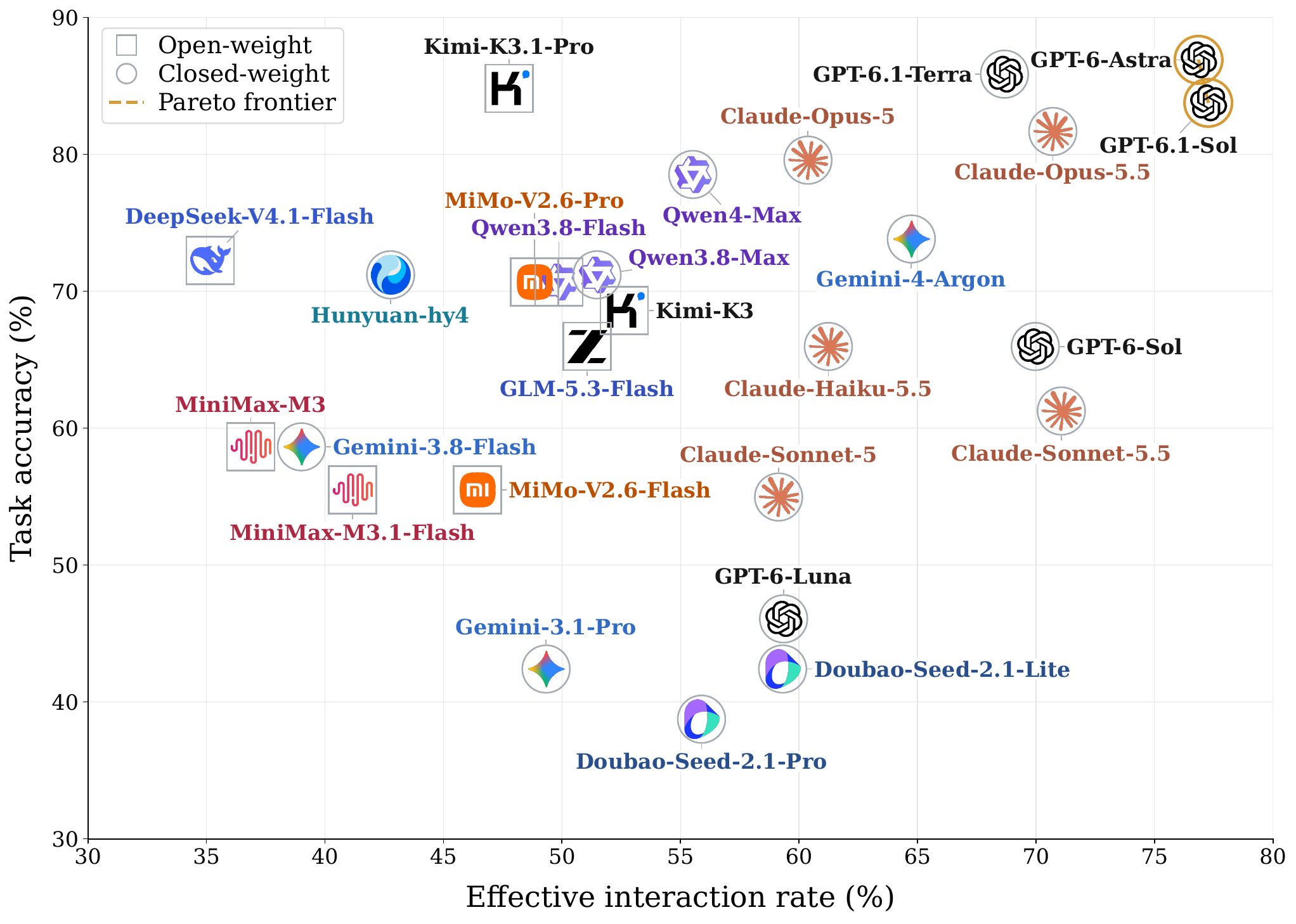}
\caption{\textbf{Accuracy vs.\ effective interaction rate on all 191 tasks.} Each marker is one model; the dashed line is the Pareto frontier.}
\label{fig:teb191-eir}\end{figure*}

\begin{figure*}[t]\centering
\includegraphics[width=\textwidth]{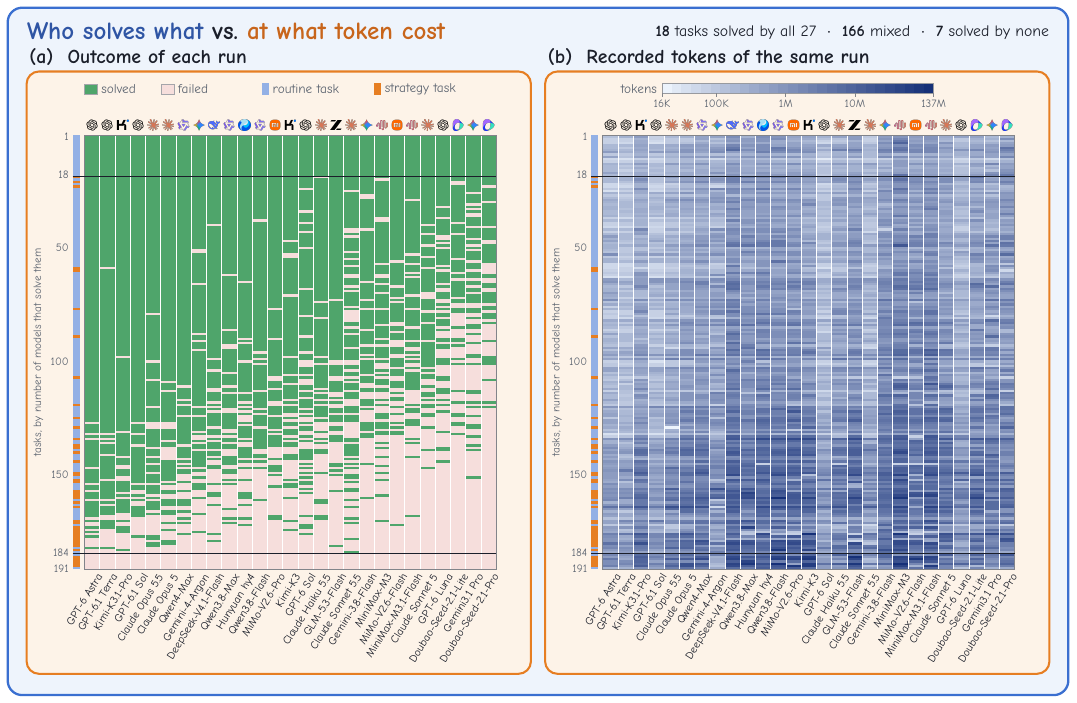}
\caption{\textbf{Outcome and recorded tokens of every selected run.} Rows are tasks, ordered by the number of models that solve them; columns are models, ordered by overall accuracy. Each cell is one run.}
\label{fig:stat-matrices}\end{figure*}

\begin{figure*}[t]
\centering
\includegraphics[width=\textwidth]{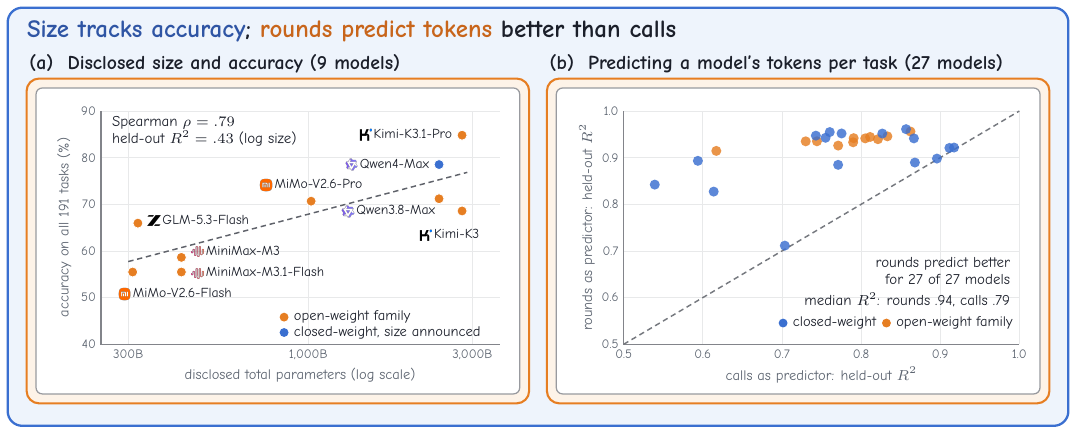}
\caption{\textbf{Size and accuracy; rounds and tokens.} (a) Accuracy on all 191 tasks against the disclosed total parameters of the nine models in Table~\ref{tab:modelwide_sizes}; the dashed line is the fit on log size and has limited held-out accuracy. (b) For each of the 27 models, held-out $R^2$ of predicting log recorded tokens per task from log rounds (vertical) or log calls (horizontal).}
\label{fig:scale_feedback}
\end{figure*}

For sources with confirmed compatible accounting, the released logged token
total is defined from five retained components:
\begin{equation}
 \begin{aligned}
 Q_{m\tau}={}&Q_{m\tau}^{\mathrm{input}}
                      +Q_{m\tau}^{\mathrm{cache\_read}}\\
                      &+Q_{m\tau}^{\mathrm{cache\_write}}
                      +Q_{m\tau}^{\mathrm{output}}\\
                      &+Q_{m\tau}^{\mathrm{reasoning}}.
 \end{aligned}
 \label{eq:app_token_components}
\end{equation}
These are source-specific usage records. They retain new input, cache
reading, cache writing, output, and reasoning accounting without replacing
the provider's definitions. Where the
relationship among reported components does not establish a comparable
total, components are retained and that total remains unavailable.

Let $\mathcal C_m^{\mathrm{tok}}$ contain included tasks with an available
total. Cumulative and per-task mean consumption are
\begin{equation}
 \begin{aligned}
 \operatorname{Tokens}_{\mathrm{total}}(m)
 &=\sum_{\tau\in\mathcal C_m^{\mathrm{tok}}}Q_{m\tau},\\
 \overline{\operatorname{Tokens}}(m)
 &=\frac{\operatorname{Tokens}_{\mathrm{total}}(m)}
 {|\mathcal C_m^{\mathrm{tok}}|}.
 \end{aligned}
 \label{eq:app_token_aggregation}
\end{equation}
Both retain successful and failed runs. Cumulative usage is reported with
its actual run coverage; the mean uses that same coverage denominator. Logged totals are not
treated as a provider-independent measure of physical computation or price.

\subsection{Metric Scopes and Matched Comparisons}

Aggregate tables retain each metric's defined scope: fixed planned tasks
for Acc; successful, label-available positive-call runs for EIR;
and included successful and failed runs with the required records for
Steps, C/T, and tokens. Missing values, empty sets, and zero ratio
denominators are reported as unavailable.

For a strict comparison of metric $g$ across models $\mathcal M$, the
common-success analysis set is
\begin{equation}
 \mathcal P_{\mathcal M}^{g}
 =\bigcap_{m\in\mathcal M}
 \{\tau\in\mathcal S_m:g(m,\tau)\text{ is available}\}.
 \label{eq:app_paired_metric_set}
\end{equation}
Such comparisons hold the completed task set fixed and report model
identities, task keys, and actual denominators. They are distinguished from
headline summaries over each model's own eligible tasks.

\subsection{Recorded Fields and Offline Recalculation}

Per-task exports retain final success, calls, turns, recorded token
components, and the effective-call fraction when available. These exports
support recomputing the reported aggregates.

Replaying a run's attribution requires access to its archived session,
execution logs, associated snapshots, final artifacts, and the matching
task assets and scoring version (Appendix~\ref{app:provenance}).

\section{Additional Experimental Analyses}
\label{app:additional_analyses}
This appendix collects the model-wide statistics, matched comparisons, and failure analyses that support Sec.~\ref*{sec:experiments}. Trajectory case studies follow in Appendix~\ref{app:case_section}; release and scoring conventions are in Appendix~\ref{app:provenance}.

\subsection{Model-Wide Statistical Analysis}
\label{app:modelwide_methods}
\begin{table}[t]
\caption{Publicly reported parameter counts for the nine models in the size analysis (B = billions). DeepSeek-V4.1-Flash is excluded because a comparable full-model count is unavailable.}
\label{tab:modelwide_sizes}
\centering\small
\setlength{\tabcolsep}{5pt}
\begin{tabular}{lrr}\toprule
Variant & Total & Active\\\midrule
\mlogo{glm}GLM-5.3-Flash~\cite{zai2026glm53flashcard} & 320B & 18B \\
\mlogo{kimi}Kimi-K3~\cite{kimi2026k3card} & 2800B & 104B \\
\mlogo{kimi}Kimi-K3.1-Pro~\cite{kimi2026procard} & 2800B & 104B \\
\mlogo{mimo}MiMo-V2.6-Flash~\cite{mimo2026flashcard} & 309B & 15B \\
\mlogo{mimo}MiMo-V2.6-Pro~\cite{mimo2026procard} & 1020B & 42B \\
\mlogo{minimax}MiniMax-M3~\cite{minimax2026m3card} & 428B & 23B \\
\mlogo{minimax}MiniMax-M3.1-Flash~\cite{minimax2026m31flashcard} & 428B & 23B \\
\mlogo{qwen}Qwen3.8-Max~\cite{qwen2026maxcard} & 2400B & 95B \\
\mlogo{qwen}Qwen4-Max~\cite{qwen2026qwen4release} & 2400B & 71B \\
\bottomrule\end{tabular}
\end{table}
\relax%
We analyze 5,157 trajectories from 27 models on 191 tasks. Pairwise analyses cover all 351 model pairs, with results reported for the full benchmark and separately for routine and strategy tasks. Each comparison uses the task outcomes and interaction annotations of the evaluated runs.

Task success and execution cost distinguish different aspects of model performance. Figure~\ref{fig:stat-efficiency} compares accuracy with mean calls, rounds, and recorded tokens, including failed runs, for each task group. Astra and Kimi-K3.1-Pro both solve 65.1\% of the strategy tasks, yet average 26.2 and 62.4 calls. Similar completion rates can therefore require substantially different amounts of external work.

Figure~\ref{fig:teb191-eir} places the 27 models by accuracy and effective interaction rate on all 191 tasks; Figure~\ref*{fig:teaser} shows the same view for the 43 strategy tasks.

Successful-task overlap reveals substantial differences in model coverage. Only 18 tasks are solved by all 27 models. We therefore compare each pair on the tasks both models solve, keeping task identity and final success fixed while retaining a broader task sample for each comparison.

Figure~\ref{fig:stat-matrices} shows every selected run. Eighteen tasks are solved by all 27 models, seven by none, and the remaining 166 have mixed outcomes; the right panel gives the recorded tokens of the same runs in the same order.

\begin{figure*}[t]
\centering
\includegraphics[width=\textwidth]{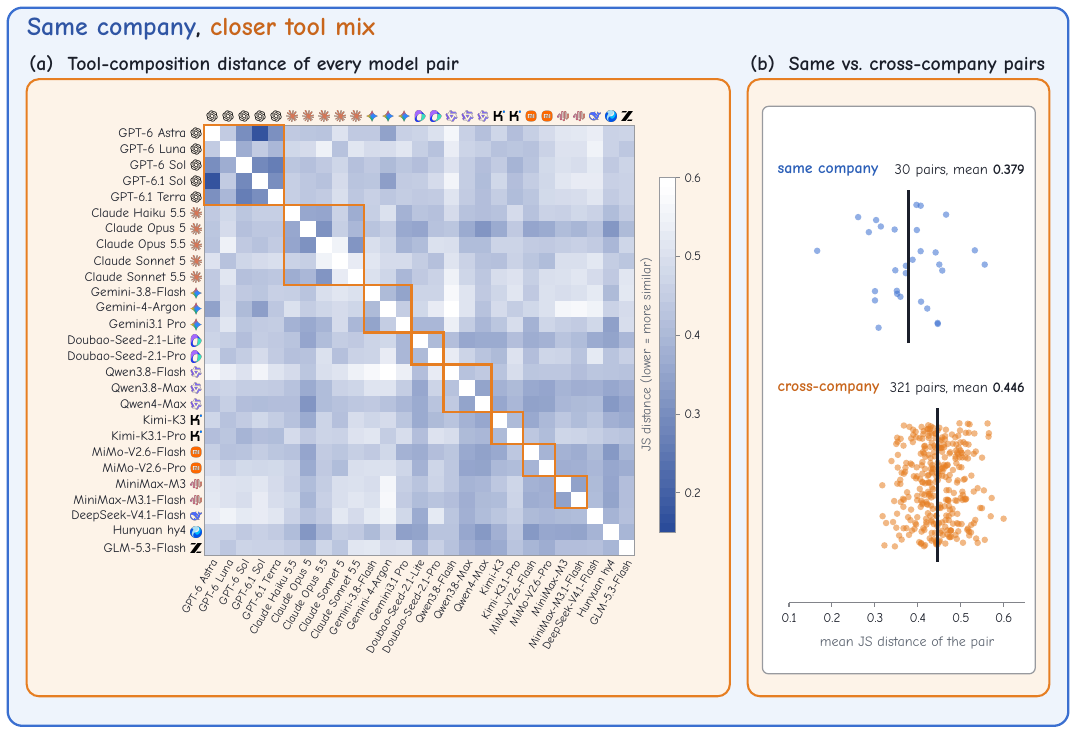}
\caption{\textbf{Same company, closer tool mix.} (a) Mean Jensen--Shannon distance between per-task tool distributions over each pair's jointly solved tasks; outlined blocks are companies. (b) The same distances by company relation; bars mark means. Pair subsets differ.}
\label{fig:family_matrix}
\end{figure*}

\subsection{Model Size and Task Adjustment}
\label{app:modelwide_scale}
The size analysis includes nine models with comparable, publicly reported total and active parameter counts (Table~\ref{tab:modelwide_sizes}). Models without comparable full-model counts, including DeepSeek-V4.1-Flash, are excluded. Total and active parameters are analyzed separately.

Larger models in the disclosed-size sample use fewer calls and rounds after adjustment for task identity. On successful runs, we fit additive task and model effects to $\log(1+\mathrm{Calls})$, $\log(1+\mathrm{Rounds})$, and other outcomes, then correlate the centered model effects with parameter counts. Total parameters correlate with adjusted calls at $\rho=-0.72$ and rounds at $-0.78$; active parameters yield $-0.71$ and $-0.77$. The call association remains negative when each company is removed in turn ($-0.72$ to $-0.57$) and when task adjustment is restricted to the nine models with disclosed sizes ($-0.67$).

Parameter count also predicts accuracy and execution compactness within this sample. An OLS accuracy regression on log total parameters has in-sample $R^2=0.677$ and leave-one-model-out $R^2=0.433$ (Figure~\ref{fig:scale_feedback}a); using untransformed total parameters gives $0.657$ and $0.393$. For the task-adjusted call effect, log size gives in-sample $R^2=0.633$ and leave-one-out $R^2=0.377$. The weaker held-out fits quantify the limits of size alone as a predictor.

\begin{figure*}[t]\centering
\includegraphics[width=\textwidth]{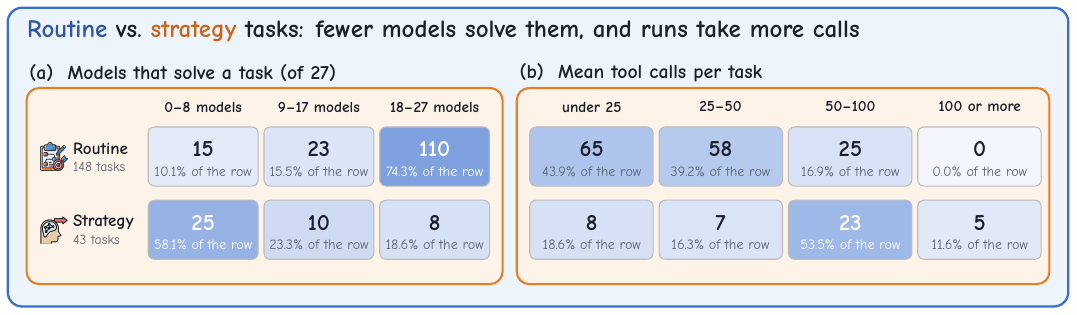}
\caption{\textbf{Routine vs.\ strategy tasks.} Tasks binned by (a) how many of the 27 models solve them and (b) their mean calls per run. Cells give task counts and row shares.}
\label{fig:stat-task-types}\end{figure*}

\begin{figure*}[t]
\centering
\includegraphics[width=\textwidth]{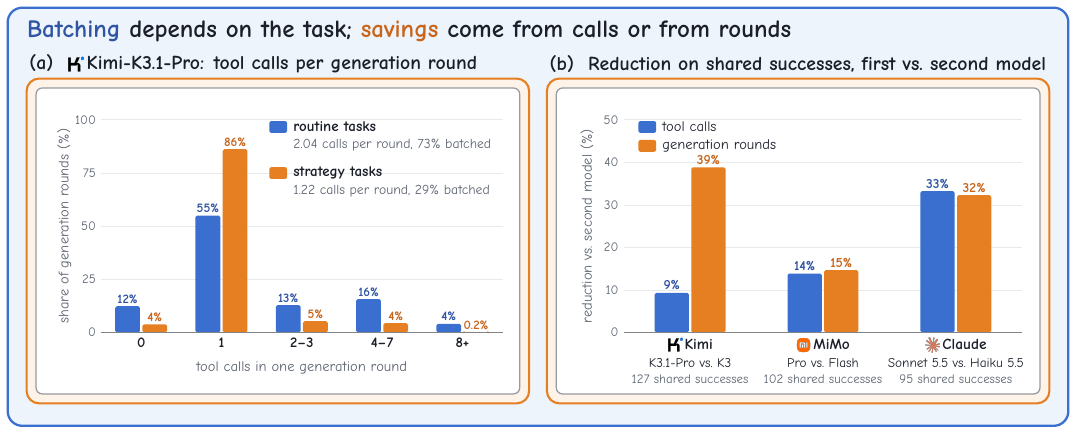}
\caption{\textbf{Batching and savings.} (a) Share of Kimi-K3.1-Pro's generation rounds by the number of tool calls they issue, over all selected runs and including rounds without calls; batched calls are those issued in rounds with more than one call. (b) Reduction in total calls and rounds of the first model relative to the second on their shared successes. Panel (b) distinguishes reductions in action volume from reductions in feedback rounds.}
\label{fig:behavior}
\end{figure*}

\begin{figure*}[t]
\centering
\includegraphics[width=\textwidth]{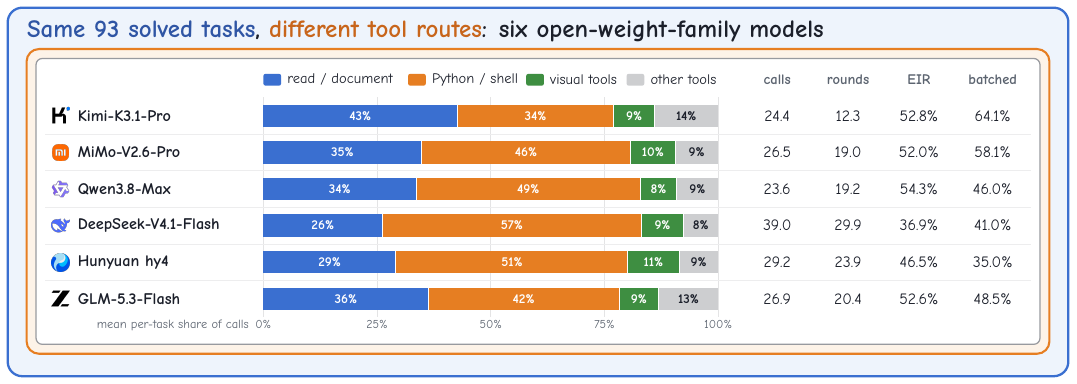}
\caption{\textbf{Same 93 solved tasks, different tool routes.} Bars average per-task shares of calls on the 93 tasks all six models solve (87 routine, six strategy). Calls, rounds, and EIR are task means; batched is the mean per-task share of calls issued in multi-call rounds. Visual tools are OCR, zoom, detection, chart extraction, barcode decoding, and image measurements.}
\label{fig:toolprofiles}
\end{figure*}

\subsection{Family Resemblance and Matched Tool Preferences}
\label{app:family_statistics}
Tool choice retains a clear family signature on jointly solved tasks. For model $m$ and task $\tau$, we normalize non-submission tool counts into a distribution $p_{m\tau}$ over tool names. For each pair, we average the base-two Jensen--Shannon distance (the square root of divergence) across its jointly successful tasks. Every pair has at least 55 such tasks. Distances give equal weight to tasks within each pair and to pairs within each comparison. Normalization separates tool choice from call volume. The 30 same-company pairs average 0.379 and the 321 cross-company pairs average 0.446 (Figure~\ref{fig:family_matrix}). Eight companies have multiple evaluated models; the other three cannot contribute a within-company pair.

Five thousand permutations of the models' family labels preserve family sample sizes. Unrestricted one-sided $p=0.0002$; permutations within weight-release groups give $p=0.0004$. Restricting further within supplier/release strata gives $p=0.0002$, the smallest attainable value. Restricting pairs to at least 75 or 100 shared successes gives within/between distances of 0.376/0.445 and 0.357/0.442, with supplier/release-stratified $p=0.0010$ and $0.0046$. Excluding Google gives 0.376/0.440 and $p=0.0016$ under the same stratification. Family resemblance therefore persists across the stratification and task-overlap checks.

For the 24 models whose companies have another evaluated model, the nearest tool-profile neighbor among these models comes from the same company in 14 cases (58.3\%); random family labels average 10.8\%. Astra and GPT-6.1 Sol are mutual nearest neighbors, as are Terra and GPT-6 Sol, and four of the five Claude models have a Claude nearest neighbor.

GPT and Claude use different tool mixes even on identical successful tasks. The ten models jointly solve 48 tasks (45 routine, three strategy), on which we average per-task tool shares within company. GPT allocates $9.94$ percentage points more of its calls to read/document and $14.54$ points less to code. Its mean call and round counts are lower by $1.72$ and $1.35$. These differences show distinct material-access and programming preferences on the predominantly routine subset.

Higher success within a family usually accompanies fewer calls and less unmatched work, but no common increase in batching. We compare the 22 model pairs with at least 30 joint successes and an overall accuracy difference of at least five percentage points. The higher-accuracy model uses fewer calls in 19 pairs, fewer rounds in 14, higher EIR in 19, and fewer unmatched calls in 19; its multi-call fraction is higher in only six. These direction counts summarize overlapping comparisons.

\subsection{Token Prediction, Task Types, and Unsuccessful Execution}
\label{app:feedback_statistics}
For model $m$ and task $\tau$, let $Q_{m\tau}$ be the recorded tokens and $J_{m\tau}$ the number of generation rounds, as in Sec.~\ref*{sec:evaluation_metrics}. We fit
\begin{equation}
 \log(1+Q_{m\tau})=a_m+b_m\log(1+J_{m\tau})+\epsilon_{m\tau}.
 \label{eq:token_round_model}
\end{equation}
Generation rounds predict recorded token usage more accurately than tool-call counts for every evaluated model. We fit Eq.~\ref{eq:token_round_model} separately for each of the 27 models on all 191 tasks, including failures, and evaluate it by leaving out one task at a time. Replacing log rounds with log calls provides the comparison model. Median held-out $R^2$ is 0.935 for rounds and 0.790 for calls (Figure~\ref{fig:scale_feedback}b); median fitted slopes are 1.67 and 1.42. Every round-based slope exceeds one, indicating that recorded tokens grow more than proportionally with rounds in the fitted relationships.

Successful strategy execution becomes more sequential across the evaluated models. All 27 models have lower strategy accuracy. Among successful runs, C/T is lower on strategy tasks for 25/27 models, and median model C/T falls from 1.31 to 1.04. The mean per-task fraction of calls issued in multi-call rounds is lower for 24/27 models, with medians of 47.4\% on routine and 18.0\% on strategy tasks.

Figure~\ref{fig:stat-task-types} gives the task-level view. Of the 43 strategy tasks, 25 are solved by at most eight models, compared with 15 of the 148 routine tasks; 28 strategy tasks average at least 50 calls per run, compared with 25 routine tasks.

Failed tasks consume a disproportionate share of recorded tokens. For every model, failed runs average more calls and rounds than successful runs, and their token share exceeds their task share. The median failure/success round ratio is 2.08. Within models, median Spearman correlations with tokens are 0.96 for rounds and 0.88 for calls; rounds correlate more strongly for 26/27 models. Appendix~\ref{app:failuretail} further separates these patterns by termination outcome.

\subsection{Additional Competence and Tool-Route Comparisons}
\label{app:tool_routes}
Across all 191 tasks, Astra, Terra, and GPT-6.1 Sol each solve 138/148 routine tasks, but solve 28, 26, and 22 of the 43 strategy tasks. Their observed separation appears in strategy tasks after their routine success counts have converged. Similar aggregate scores can also accompany different competence profiles across families. MiMo Pro and Qwen3.8-Max have nearly equal overall success (70.7\% and 71.2\%), yet MiMo Pro performs better on routine tasks (83.8\% versus 79.7\%), while Qwen3.8-Max performs better on strategy tasks (41.9\% versus 25.6\%). These models exhibit different strengths in prescribed execution and strategy construction on this task set.

Qwen3.8-Max and Hunyuan each solve 136 tasks, with 117 shared successes and 19 exclusive successes each. On the 117 shared tasks, Qwen3.8-Max uses 25.03 calls versus 34.11 for Hunyuan, while EIR is 53.43\% versus 44.45\%. Thus models can differ in both task specialization and the amount of external action needed for tasks they both solve.

Among the Gemini models, Argon and Flash jointly solve 105 tasks. Argon uses 16.43 versus 35.95 calls, with EIR of 68.02\% versus 39.10\% (Table~\ref*{tab:paired}). For Qwen4-Max and Qwen3.8-Max on their 127 shared successes, the round difference is $-2.10$ and the EIR difference is $+4.02$ percentage points.

\begin{table*}[t]
\caption{Disclosed scale and execution on shared successes drawn from the full 191-task pool. Parameter pairs are larger / smaller; B and T denote billions and trillions. Call changes and EIR differences are larger minus smaller on identical jointly successful tasks. Sizes are publicly reported architecture counts.}
\label{tab:size}
\centering\small
\setlength{\tabcolsep}{6pt}
\begin{tabular}{lllrrr}\toprule
Larger / smaller variant & Total parameters & Active parameters & Shared & $\Delta$ Calls (\%) & $\Delta$ EIR (pp)\\\midrule
\begin{tabular}[c]{@{}l@{}}\mlogo{mimo}MiMo-V2.6-Pro~\cite{mimo2026procard}\\\mlogo{mimo}MiMo-V2.6-Flash~\cite{mimo2026flashcard}\end{tabular} & 1.02T / 309B & 42B / 15B & 102 & $-13.9$ & $+4.92$\\
\bottomrule
\end{tabular}
\end{table*}

\begin{table*}[t]
\centering\small
\setlength{\tabcolsep}{6pt}
\begin{tabular}{lrrr}\toprule
Per-task measurement & \mlogo{kimi}Kimi-K3.1-Pro~\cite{kimi2026procard} & \mlogo{kimi}Kimi-K3~\cite{kimi2026k3card} & Difference\\\midrule
\tebtoolicon{read}Read requests to initial-image paths & 3.29 & 0.43 & $+2.86$\\
\tebtoolicon{zoom}Zoom calls & 2.96 & 7.67 & $-4.71$\\
\tebtoolicon{python}Python calls & 4.55 & 10.57 & $-6.02$\\
\tebtoolicon{ocr}OCR calls & 2.73 & 2.73 & $0.00$\\
\phantom{\tebtoolicon{read}}All calls & 27.55 & 33.65 & $-6.10$\\
\phantom{\tebtoolicon{read}}Generation rounds & 13.82 & 25.84 & $-12.02$\\\bottomrule
\end{tabular}
\caption{Observation routes on the same 49 jointly solved tasks supplying raster images. Differences are Pro minus K3. Read measurements count requests to supplied paths.}
\label{tab:visualroutes}
\end{table*}

\subsubsection{Task-Dependent Call Organization}
\label{app:rounds_extra}
On the 151 tasks that Astra, Terra, and GPT-6.1 Sol all solve within the full 191-task set, calls average 14.68, 18.14, and 14.38, while generation rounds average 15.81, 9.66, and 13.60. Mean per-task fractions of calls in multi-call rounds are 2.8\%, 57.9\%, and 15.8\%. Terra places more external operations within fewer generations. Astra nevertheless invokes Python more often than Terra (2.55 versus 1.72 calls), while using fewer document calls (4.25 versus 5.75) and shell calls (0.99 versus 2.20). This localizes part of Astra's call savings to document and shell access, while Python use increases (Appendix~\ref{app:newprofiles}).

Kimi-K3.1-Pro shows pronounced task-dependent batching. Its C/T is 2.04 on routine tasks and 1.22 on strategy tasks, with mean rounds increasing from 16.09 to 51.33. The median nonempty batch is one in both groups; larger batches account for 73.1\% of calls on routine tasks but 29.0\% on strategy tasks (Figure~\ref{fig:behavior}a). In a successful receipt audit, successive eight-call rounds apply OCR to eight different receipt images, illustrating broad acquisition of multiple pieces of evidence within one generation.

The shift toward lower C/T extends across models. Restricting to successful runs, 25 of the 27 models have lower C/T on strategy tasks. Median model C/T changes from 1.31 to 1.04. Successful strategy execution therefore tends toward one call per generation even among models that batch more heavily on routine tasks. This is consistent with more successive observation and decision cycles. Batching here describes emitted calls, not measured backend concurrency.

\subsubsection{Six-Model Tool Profiles}
To compare tool use on identical successful tasks, we select the 93 tasks all six models in Figure~\ref{fig:toolprofiles} solve. This subset contains 87 routine and six strategy tasks. The comparison therefore primarily characterizes routine execution.

K3.1 Pro has a larger mean per-task read/document share (42.7\%) than code share (34.4\%) and batches more heavily. Qwen3.8-Max uses a larger code share (49.3\%), mixing shell and Python. MiMo Pro combines document access with programmatic operations; GLM Flash has a read/document share of 36.3\% and code share of 41.9\%. Hunyuan uses 9.02 Python calls per task versus Qwen3.8-Max's 5.03. DeepSeek Flash uses a code share of 57.1\%, with a longer mean maximum chain of consecutive pure-code rounds (11.87 versus Qwen3.8-Max's 7.51). A pure-code round contains Python or shell calls and no other non-submit tools; the chain statistic is the task mean of each run's longest consecutive chain of such rounds.

Distinct routes can yield similar point estimates of efficiency: K3.1 Pro and Qwen3.8-Max average 24.42 and 23.58 calls, with EIRs of 52.8\% and 54.3\%, while using 12.3 and 19.2 rounds. Thus batching and tool preference characterize different ways of accomplishing these tasks.

Claude Opus 5.5, Sonnet 5.5, and Haiku 5.5 jointly solve 91 benchmark tasks. Opus and Sonnet average 14.46 and 13.81 calls, versus Haiku's 20.76; EIR is 76.5\%, 75.3\%, and 64.0\%. Haiku uses more Python (5.52 versus Opus's 3.00) and document calls (6.31 versus 4.08). Opus and Sonnet have similar call means on this subset. Opus's higher overall accuracy is expressed in the number of tasks solved, while their shared-task execution lengths remain close. Full profiles appear in Appendix~\ref{app:newprofiles}.

\begin{figure*}[t]
\centering
\includegraphics[width=\textwidth]{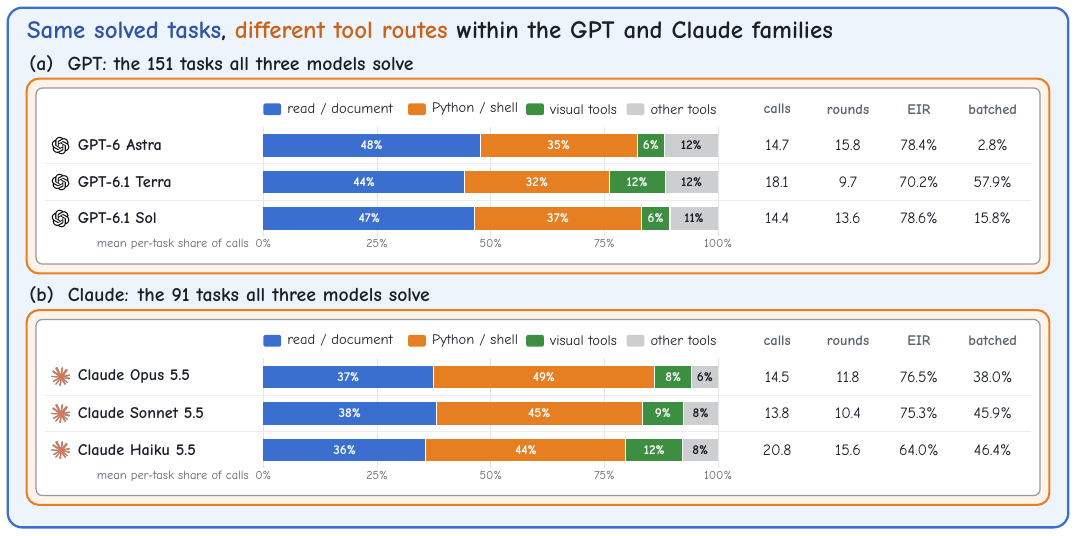}
\caption{\textbf{Tool routes within the GPT and Claude families.} Each panel uses the tasks all three of its models solve, so the two panels cover different tasks. Columns and tool groups as in Figure~\ref{fig:toolprofiles}; batched is the emitted multi-call share, not measured concurrency.}
\label{fig:newprofiles}
\end{figure*}

\begin{figure*}[t]\centering
\includegraphics[width=\textwidth]{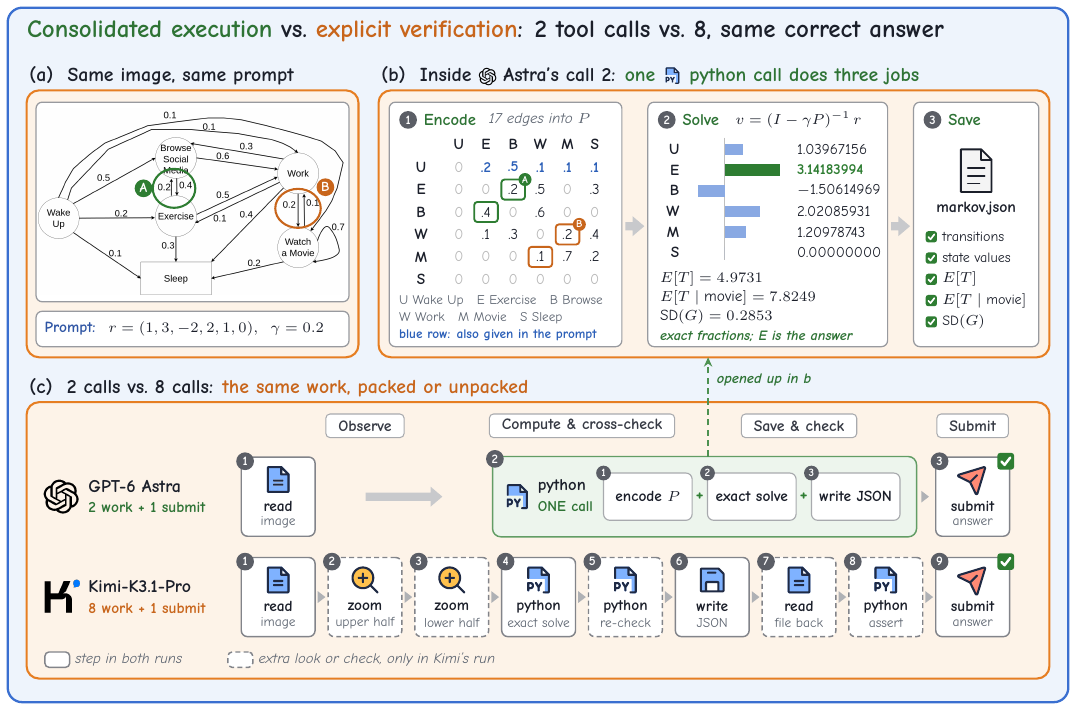}
\caption{\textbf{Consolidated execution vs.\ explicit verification.} GPT-6 Astra folds encoding, solving, and saving into one Python call; Kimi-K3.1-Pro spreads the same work over eight calls with extra looks and checks. Both answers are correct.}
\label{fig:case-markov}\end{figure*}

\begin{figure*}[t]\centering
\includegraphics[width=\textwidth]{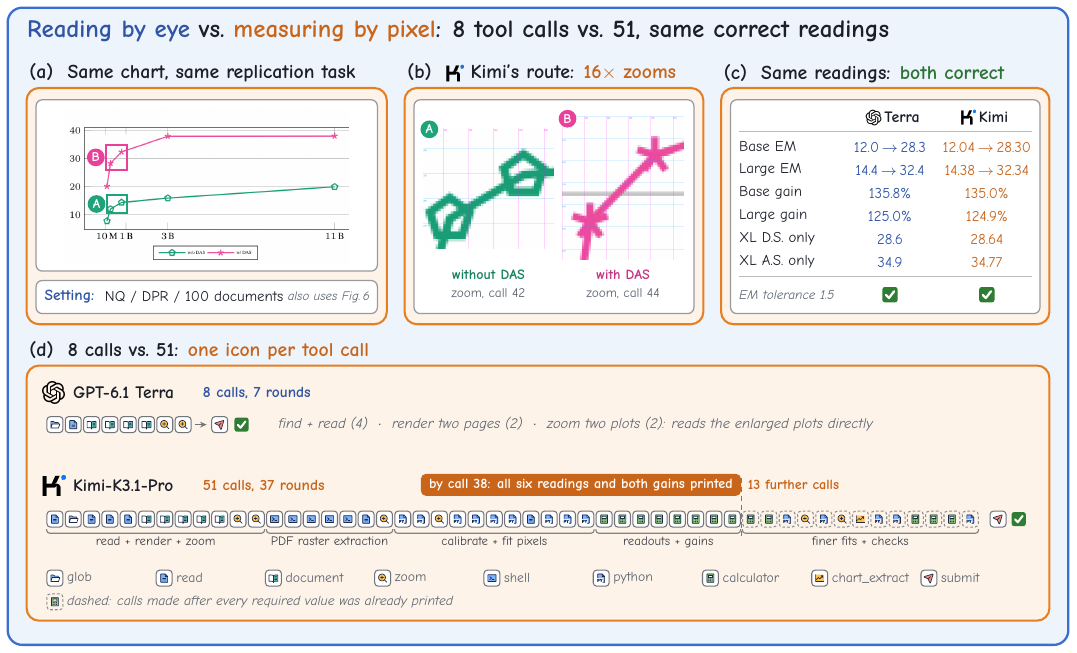}
\caption{\textbf{Reading by eye vs.\ measuring by pixel.} GPT-6.1 Terra reads two enlarged plots; Kimi-K3.1-Pro calibrates axes and fits pixels, and keeps refining after every required value is printed. Both sets of readings are correct.}
\label{fig:case-reader}\end{figure*}

\begin{figure*}[t]\centering
\includegraphics[width=\textwidth]{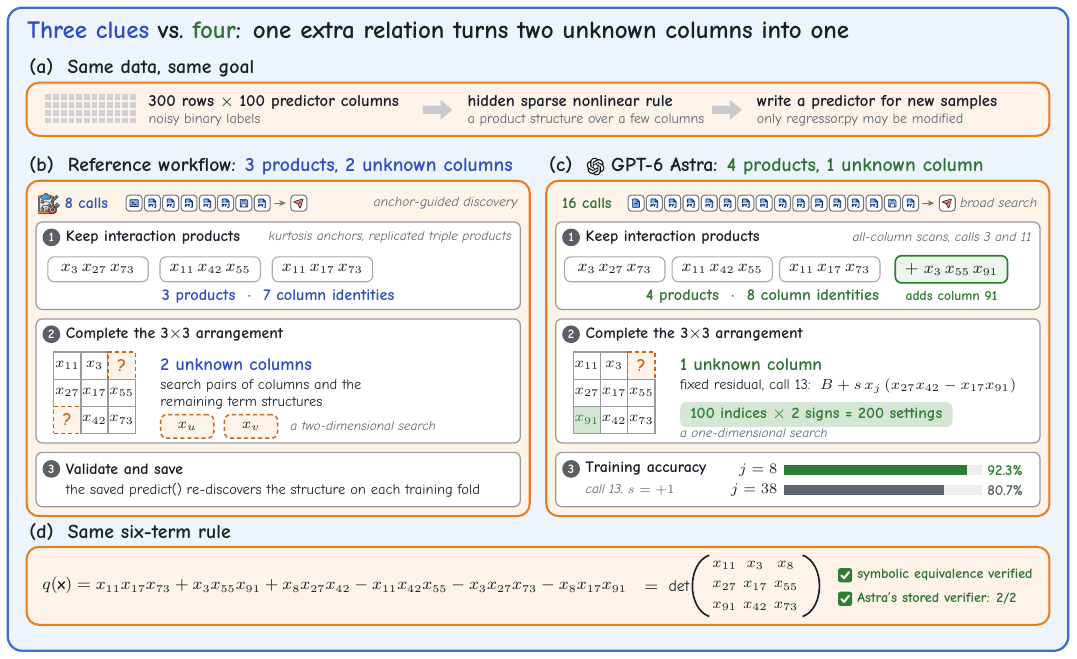}
\caption{\textbf{One extra intermediate relation reshapes the search.} The reference workflow keeps three products and searches two unknown columns jointly; GPT-6 Astra keeps a fourth, leaving a one-dimensional search. Both recover the same rule.}
\label{fig:case-symbolic}\end{figure*}

\begin{figure*}[p]\centering
\includegraphics[width=.92\textwidth]{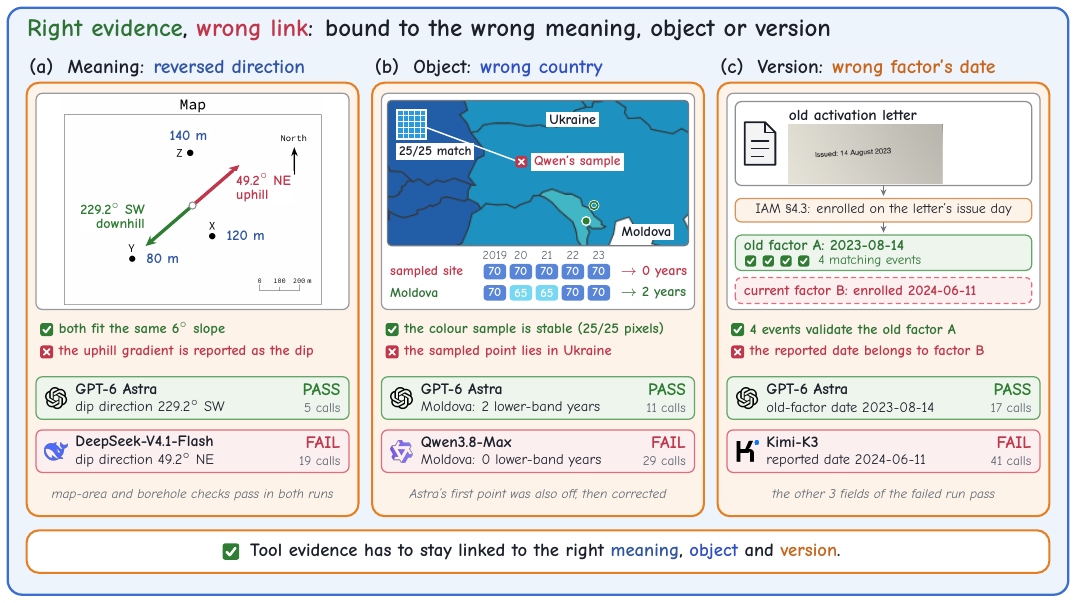}
\caption{\textbf{Right evidence, wrong link.} In three failed runs the tool evidence is sound but is bound to the wrong meaning (uphill reported as dip), object (a stable sample in the neighboring country), or version (the current factor's date instead of the validated old one).}
\label{fig:case-links}
\vspace{1em}
\includegraphics[width=.92\textwidth]{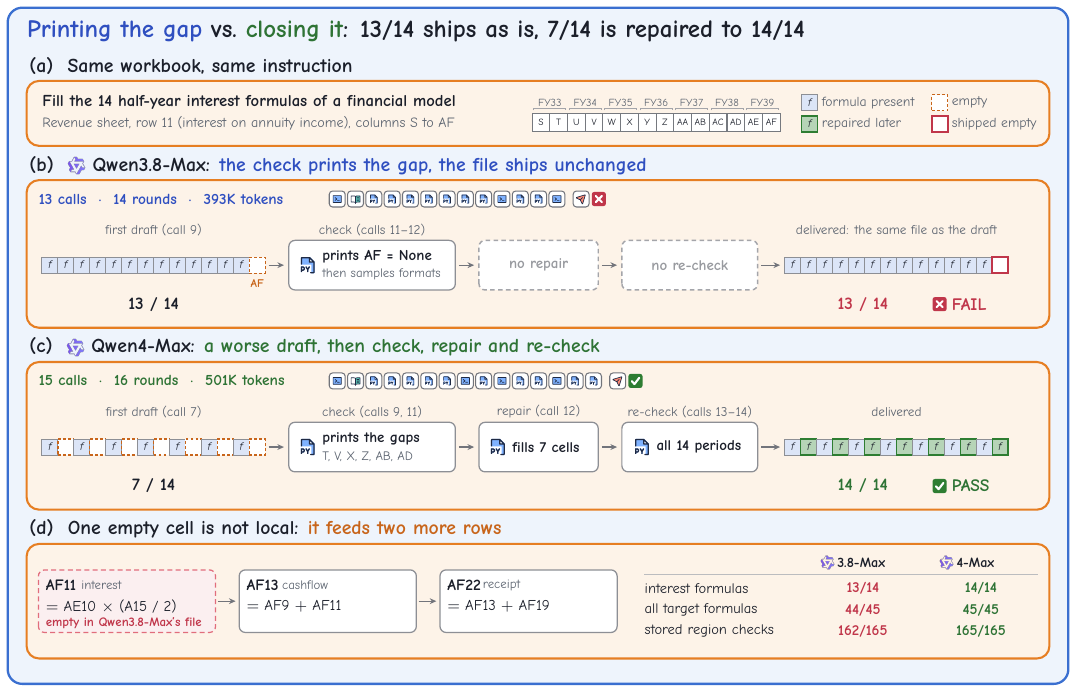}
\caption{\textbf{Printing the gap vs.\ closing it.} Qwen3.8-Max prints the one empty formula cell and ships the file unchanged; Qwen4-Max starts from a worse draft, finds the gaps, repairs them, and re-checks every period.}
\label{fig:case-workbook}
\end{figure*}
\subsubsection{Operation Routes and Evidence-Label Savings}
\begin{table}[t]
\centering\small
\setlength{\tabcolsep}{3.5pt}
\begin{tabular}{lrrr}\toprule
Pair & \shortstack{Calls\\saved} & \shortstack{Unmatched\\share} & \shortstack{Repeat +\\error share}\\\midrule
\begin{tabular}[c]{@{}l@{}}\mlogo{mimo}MiMo-V2.6-Pro~\cite{mimo2026technicalreport}\\\mlogo{mimo}MiMo-V2.6-Flash~\cite{mimo2026technicalreport}\end{tabular} & 4.60 & 82.3\% & 7.7\%\\
\begin{tabular}[c]{@{}l@{}}\mlogo{qwen}Qwen3.8-Max~\cite{qwen2026maxcard}\\\mlogo{hunyuan}Hunyuan hy4~\cite{tencent2026hy4release}\end{tabular} & 9.08 & 83.6\% & 0.9\%\\
\bottomrule
\end{tabular}
\caption{Composition of the net decrease in calls on shared successes drawn from the full 191-task pool ($n=102,117$). Savings are the second model minus the first. Shares use released evidence labels: unmatched means the non-repeat redundant category. The remainder is a decrease in matched-effective calls. These are differences in mean counts, not fractions of individual runs.}
\label{tab:saved}
\end{table}

MiMo Pro's savings concentrate in programmatic interaction. On its 102 shared successes with Flash, Python calls decrease from 8.22 to 4.98, while combined read/document calls remain similar (10.42 versus 10.43). The mean longest chain of consecutive pure-code rounds falls from 7.41 to 5.85. Pro's reduction therefore appears mainly in the code-interaction path, alongside its higher EIR. Matched-effective call counts remain similar (15.23 versus 14.76), while unmatched calls decrease substantially. Table~\ref{tab:size} compares the reported architecture sizes and execution statistics of the two models on these tasks.

K3.1 Pro changes its observation route. On 127 shared successes with K3, Python calls decrease from 7.24 to 4.50, zoom from 3.40 to 1.39, and read increases from 3.16 to 5.22. Image-file reads account for much of the read increase, rising from 0.46 to 1.77 calls per task. Document calls remain similar (8.07 versus 7.96). This describes more direct image reading alongside fewer zoom and code calls. Calculator calls increase from 0.26 to 0.61.

The route shift also appears on a subset defined by task materials. The two Kimi models jointly solve 49 tasks that supply raster images (43 routine, six strategy). Pro makes 3.29 read requests to original-image paths per task versus K3's 0.43, while zoom falls from 7.67 to 2.96 and Python from 10.57 to 4.55. OCR remains 2.73 for both. Mean total calls fall from 33.65 to 27.55 and rounds from 25.84 to 13.82. Table~\ref{tab:visualroutes} lists these measurements. This localizes the observation-route change to tasks supplying images. This shift is consistent with the stronger visual benchmark results reported for K3.1 Pro: PerceptionBench rises from 58.5 to 68.3 and MMMU-Pro from 81.6 to 84.1~\cite{kimi2026k3card,kimi2026procard}.

The read/zoom change includes a substitution between image-access interfaces. On these 49 tasks, native reads plus successful whole-image zoom returns average 3.71 versus 3.63 per task, so whole-image access remains nearly unchanged. In the selected wind/solar task, K3 repeatedly requests zoom without a bounding box before switching to a Python crop, whereas Pro takes a shorter route. The example locates a concrete difference in parameter selection and recovery (Appendix~\ref{app:cases}).

The MiMo call advantage is not confined to the most favorable cases. After removing the 5\% of shared-success tasks with the largest call savings, MiMo Pro still saves 2.92 calls relative to Flash. For Kimi, the same removal leaves a difference of $-0.57$ calls; its main observed change is in feedback-round organization. Both shared-success subsets are predominantly routine: 95/102 tasks for MiMo and 114/127 for Kimi.

The evidence annotations locate where compact execution differs (Table~\ref{tab:saved}). Of MiMo Pro's 4.60 fewer calls relative to Flash, 3.78 are in the unmatched redundant category, accounting for 82.3\% of the net decrease. Repeat and error reductions together account for 7.7\%. For Qwen3.8-Max versus Hunyuan, unmatched calls explain 83.6\% of the 9.08-call decrease, while repeat and error reductions explain 0.9\%.

Most saved calls concern intermediate work without new matched evidence, while exact repetitions and tool errors explain much less of the gap. Action selection and task decomposition are therefore the main observed behavioral distinctions behind compact execution in these paired comparisons.

\subsection{Full-Task GPT and Claude Tool Profiles}
\label{app:newprofiles}
Tool profiles are compared on tasks jointly solved by three models within each family: GPT-6 Astra, GPT-6.1 Terra, and GPT-6.1 Sol; and Claude Opus 5.5, Sonnet 5.5, and Haiku 5.5 (Figure~\ref{fig:newprofiles}). The GPT subset contains 151 tasks (133 routine, 18 strategy), and the Claude subset contains 91 (83 routine, eight strategy). We count all tool calls except answer submission, including failed calls. Batch share is the per-task fraction of calls issued in rounds containing multiple calls, averaged over tasks. Each family is therefore compared on its own successful-task intersection.

On the GPT subset, Astra minus Terra differences are $+0.83$ Python calls, $-1.21$ shell calls, and $-1.51$ document calls. Opus minus Haiku on the Claude subset has $-2.52$ Python calls and $-2.23$ document calls. Opus minus Sonnet has $+0.65$ total calls.

\subsection{Observation Routes on Tasks Supplying Raster Images}
\label{app:visualroutes}
Seventy-five tasks provide raster images as input. Of these, 49 are jointly solved by Kimi-K3.1-Pro and K3 (43 routine, six strategy). Image access is measured by explicit read requests to supplied image files. Combining native reads with successful zoom calls that return an entire supplied image gives 182 versus 178 accesses, or 3.71 versus 3.63 per task. The mean difference of 0.08 shows that whole-image access remains nearly unchanged as the preferred interface shifts.

\subsection{Matched Token Components}
\label{app:tokens}
The full 191-task set yields 158 joint successes for Astra/Terra, 118 for Opus/Haiku, and 95 for Sonnet/Haiku. On Astra/Terra shared successes, mean cache reads are 320,762 versus 251,317 tokens, a difference of 69,445; reported reasoning is 693 versus 1,049 tokens, a difference of $-356$. The total-token difference is 65,168. Astra's higher recorded total is mainly reflected in cache reads, despite its lower reported reasoning component. The cache-read difference exceeds the total difference because the remaining components partly offset it.

Opus/Haiku shared successes average 13.62/20.02 calls, 11.46/15.18 rounds, and 433,533/839,210 tokens. Sonnet/Haiku shared successes average 14.43/21.63 calls, 10.85/16.03 rounds, and 418,704/898,875 tokens. Their mean token differences are $-405{,}677$ and $-480{,}171$.
\subsection{Unsuccessful-Execution Analysis}
\label{app:failuretail}
The shorter execution of Kimi-K3.1-Pro relative to K3 also appears on tasks both models fail. On these 25 tasks, K3.1 Pro uses 73.72 calls and 52.84 rounds versus K3's 114.12 calls and 101.32 rounds, with recorded mean tokens of 5.31M versus 20.09M. Restricting further to the 16 tasks where both submit an incorrect answer preserves the round reduction (52.44 versus 99.69). The pattern therefore remains when these runs share the same recorded termination category.

Nevertheless, unresolved tasks continue to generate long interactions. K3.1 Pro's failed runs average 50.93 rounds, versus 19.21 for its successful runs; the corresponding K3 means are 76.35 and 25.18. K3.1 Pro spends 44.3\% of recorded tokens on its 15.2\% failed tasks; K3 spends 77.0\% on its 31.4\% failed tasks. Calls and total tokens correlate within these models (Spearman $\rho=0.75$ and $0.91$). Repeatedly processing growing histories can amplify this resource tail. Unsuccessful execution therefore accounts for a major part of resource consumption, particularly for K3.

MiMo Pro's 52 shared failures with Flash average 62.19 versus 44.48 rounds. This difference shrinks to 51.04 versus 49.58 on the 26 runs with the same recorded termination category, and reverses on the eight runs where both submit incorrect answers (86.88 versus 96.50). The aggregate failure-length gap is therefore sensitive to how the runs terminate.

\section{Trajectory Case Studies}
\label{app:case_section}
\subsection{Selected Trajectory Comparisons}
\label{app:case_studies}

Figures~\ref{fig:case-markov}--\ref{fig:case-workbook} complement Figure~\ref*{fig:case-polyhedron} with five further selected comparisons. Two contrast successful runs with different call counts: one Python call can carry several steps (Figure~\ref{fig:case-markov}), and finer measurement keeps adding calls after every required value is available (Figure~\ref{fig:case-reader}). Figure~\ref{fig:case-symbolic} compares a stored reference workflow with Astra on a discovery task. Figure~\ref{fig:case-links} collects three failures in which sound tool evidence is attached to the wrong meaning, object, or version. Figure~\ref{fig:case-workbook} contrasts a check whose finding is printed with one that is followed by repair and re-checking. All figures show selected published runs and count calls without the submission; the paired runs differ in providers and batches, so they illustrate mechanisms and do not rank models.

\subsection{Illustrative Operation-Scope Cases}
\label{app:cases}
A selected case illustrates how operation parameters change execution; it was chosen for its large difference and illustrates a possible route rather than its population frequency. In a wind/solar example, Kimi-K3.1-Pro uses 29 calls and K3 uses 70. K3 makes 19 zoom requests without bounding boxes in steps 49--67; 11 fail on the large source image and eight return whole images. It then crops through Python and reads the crop. The requests expose an operation-parameter and recovery issue. A counterexample is a successful nutrition-label task where Pro uses 76 calls versus K3's 42, including more OCR and document access.

\section{Release Provenance and Reproducibility}
\label{app:provenance}
The main leaderboard covers all 191 tasks (148 routine, 43 strategic) and reports 5,157 selected runs for 27 models, which total 194,357 non-submit calls and 165,033 generation rounds. Behavioral analyses draw their stated task subsets from this pool.

\noindent\textbf{Release.}
The public release consists of the benchmark task packages, with their checkers and evidence graphs, and the execution environment. Code and documentation are available at \url{https://github.com/SeerRay-Lab/TAE-Bench}; task packages and tool-model weights are available at \url{https://huggingface.co/datasets/SeerRay-Lab/TAE-Bench}.

\noindent\textbf{Execution and scoring.}\label{app:policy}
All models run through the shared \texttt{opencode}~\cite{opencode2026} harness and tool interface. One selected run per model and task enters the leaderboard; selection follows fixed rules and does not pick the highest-scoring attempt. Reported scores and EIR values retain their source evaluations on the frozen task versions. Token totals sum the five recorded components of the selected runs; because providers account for them differently, totals are not interpreted as internal computation.

\end{document}